\documentclass[lettersize,journal]{IEEEtran}
\usepackage{amsmath,amsfonts}
\usepackage{amsthm}
\usepackage{algorithmic}
\usepackage{algorithm}
\usepackage{array}
\usepackage{textcomp}
\usepackage{stfloats}
\usepackage{url}
\usepackage{verbatim}
\usepackage{graphicx}
\usepackage{cite}
\usepackage{multirow}
\usepackage{booktabs}
\usepackage{hyperref}
\usepackage{bm}
\usepackage{stmaryrd}
\usepackage{amssymb}
\usepackage{pifont}
\usepackage{subcaption}
\usepackage{boldline}

\newcommand{\rac}{{\sc RACLearn}}
\newcommand{\pro}{{\sc Predict-Repair-Optimize}}

\newcommand{\cmark}{\ding{51}}%
\newcommand{\xmark}{\ding{55}}%
\def\naive{na\"{i}ve}
\newcommand{\x}{\mathbf{x}}
\newcommand{\y}{\mathbf{y}}
\usepackage{enumitem,kantlipsum}

\newtheorem{theorem}{Theorem}

\begin{document}

\title{Confidence-Aware Graph Neural Networks for Learning Reliability Assessment Commitments}

\author{
    Seonho Park, Wenbo Chen, Dahye Han, Mathieu Tanneau, and Pascal Van Hentenryck
    \thanks{
    The authors are affiliated with the School of Industrial and Systems Engineering, Georgia Institute of Technology, Atlanta, GA 30332, USA, E-mail: \{seonho.park, wenbo.chen, dahye.han, mathieu.tanneau\}@gatech.edu, pascal.vanhentenryck@isye.gatech.edu
    }
}

\maketitle
\begin{abstract}
Reliability Assessment Commitment (RAC) Optimization is increasingly
important in grid operations due to larger shares of renewable
generations in the generation mix and increased prediction errors.
Independent System Operators (ISOs) also aim at using finer time
granularities, longer time horizons, and possibly stochastic
formulations for additional economic and reliability benefits. The
goal of this paper is to address the computational challenges arising
in extending the scope of RAC formulations. It presents \rac{} that 
(1) uses a Graph Neural Network 
(GNN) based architecture to predict generator commitments and active line constraints, 
(2) associates a confidence value to each commitment prediction, (3)
selects a subset of the high-confidence predictions, which are (4)
repaired for feasibility, and (5) seeds a state-of-the-art
optimization algorithm with feasible predictions and active
constraints. Experimental results on exact RAC formulations used
by the Midcontinent Independent System Operator (MISO) and an actual
transmission network (8965 transmission lines, 6708 buses, 1890
generators, and 6262 load units) show that the \rac{} framework
can speed up RAC optimization by factors ranging from 2 to 4 with
negligible loss in solution quality. 
\end{abstract}


\section{Introduction}
\label{sec:introduction}
\subsection{Background and Motivation}

The Security-Constrained Unit Commitment (SCUC) problem is a critical
tool to ensure reliable and economical scheduling of energy
production. Its reliability is however being challenged by the rapid
transition from fossil-fueled generators to renewable energy sources
and the policies supporting/accelerating this migration. California,
for instance, aims at having 100\% of the state's electricity procured
and served using renewable energy by 2045 \cite{CEC}. Integrating more
renewable energy sources into the grid creates significant volatility,
both in front and behind the meter, and increased forecasting errors
in net load (load minus renewable generation). For these reasons,
Independent System Operators (ISOs) use various Reliability Assessment
Commitment (RAC) models\footnote{The RAC is called the Reliability Unit Commitment (RUC) by some ISOs} \cite{ma2009security,chen2012resource,chen2014applying,BPM_C} to secure sufficient resources in the
day-ahead (DA) and real-time markets. For instance, MISO, the
Midcontinent Independent System Operator, uses both Day-Ahead Forward
Reliability Assessment Commitment (DA-FRAC) and Look-Ahead
Commitment (LAC) for reliability assessment. The DA-FRAC is executed
after the DA market, replaces demand bids by
load forecasts, and decides additional commitments of the
generators abiding by the DA market decisions. The LAC is executed every
15 minutes on a rolling basis and provides an additional chance to
turn on generators for a 3-hour look-ahead window. Both DA-FRAC and
LAC exploit updated forecasts for load and renewable generation.

ISOs have been striving to curtail the execution
times of the SCUC, FRAC, and LAC models in order to obtain additional
safety and economic gains. This includes the opportunity to have finer
time granularity and longer horizons in the model formulations and
to explore scenario-based approaches. For example, MISO has decreased
the DA market-clearing time from 4 hours to 3 hours in 2016
\cite{chen2016improving} and is committed to decreasing it
further.

The RAC formulations are Mixed Integer
Linear Programming (MILP) that are computationally
challenging. They are operated by MISO to deal with significant
changes in load and renewable energy forecasts. A RAC
optimization helps operators determine which additional generators
must be committed to taking into account the uncertainty in forecasting. Even
though the number of variables is smaller than in the SCUC (because
must-run commitments have already been determined by the day-ahead
or prior RAC commitments), the RAC formulations must give operators
reliable solutions to different scenarios as fast as possible, which
makes a machine-learning (ML) approach particularly appealing.
Fortunately, the FRAC and RAC are often run on similar
instances. Indeed, the pool of generators and the topology of the
network evolve slowly, while the demand and renewable generation lead
to instances that exhibit strong correlations and similar
patterns. This presents a significant opportunity for ML techniques
that can then be used to speed up the solution process of RAC
formulations.

This is precisely the goal of this paper, which proposes the \rac{}
framework for accelerating the RAC optimizations. 
\rac{} consists of three key components: 
(1) a Graph Neural Network (GNN) based architecture that predicts both the generator commitments and the active transmission constraints;
(2) the use of an epistemic uncertainty measurement to associate 
a confidence value with each commitment prediction; and
(3) a feasibility restoration that transforms possibly infeasible 
instances with fixed commitment variables into feasible ones in 
polynomial time. Note that the proposed \rac{} framework consolidates all those components using end-to-end learning for accelerating the RAC optimization.

\noindent
The contributions of the paper can be summarized as follows:
\begin{enumerate}[leftmargin=0.5cm]

\item{\em The \rac{} framework:} The paper presents an {\em
  end-to-end} optimization learning approach to obtain near-optimal 
  solutions to industrial RAC formulations and reduce solving times 
  by factors from 2 to 4 on an actual grid.
  
\item{\em The GNN-based architecture:} \rac{} includes 
  a novel GNN-based architecture to predict generator commitments and 
  active transmission constraints simultaneously. 
  The GNN-based architecture includes novel
  encoder-decoder layers leading to a compact design with a small
  number of trainable parameters, which is critical to scale to
  realistic networks. Experimental results show that the proposed
  architecture provides significant improvements compared to deep and
  shallow baselines.

\item{\em The confidence measurement:} \rac{} uses confidence measurements
  to select a subset of high-confidence commitment predictions. 
  Experimental results show that the confidence values
  are strongly indicative of being accurate commitment estimates. 

\item{\em The polynomial-time feasibility restoration step:}
  \rac{} includes a polynomial-time
  feasibility restoration optimization that transforms possibly 
  infeasible instances
  into feasible instances that will lead to near-optimal solutions. 
\end{enumerate}

\noindent
The rest of the paper is organized as follows.
Section~\ref{sec:related} reviews prior related works.
Section~\ref{sec:preliminaries} revisits RAC formulations and their
solution techniques. Section~\ref{sec:overview} gives an overview of the 
\rac{} framework. Section~\ref{sec:architecture} presents the proposed
GNN-based architecture. Section~\ref{sec:confidence} describes how to
compute the confidence measurements with the GNN-based architecture.
Section~\ref{sec:optimization_speedup} introduces the
\pro{} heuristic for accelerating RAC. 
Section~\ref{sec:experiments} reports the experimental settings and 
results.
Section~\ref{sec:conclusion} contains the concluding remarks.

\section{Related Work}
\label{sec:related}
Solving SCUC and RAC formulation is computationally challenging and
alternative MIP formulations of SCUCs have been developed to improve
efficiency; they include tightening the constraint polytope (e.g.,
\cite{pan2016polyhedral,damci2016polyhedral,ostrowski2011tight}),
constraints elimination (e.g.,
\cite{chen2016improving,xavier2019transmission})
and decomposition method (e.g.,
\cite{kim2018temporal,feizollahi2015large}).

The RAC formulations, which include the DA-FRAC and LAC optimization, 
have been developed and operated by MISO 
\cite{BPM_C,ma2009security,chen2012resource,chen2014applying}. 
However, to the best of the authors' knowledge, ML schemes to
accelerate the RAC optimization process have not yet been studied. 
Instead, in recent years, ML techniques have been proposed to
approximate market-clearing algorithms and/or to accelerate their
underlying optimization processes in general \cite{xavier2021learning,ramesh2022feasibility,tang2023graph,shekeew2023learning}.

For speeding up the SCUC, Xavier \textit{et al.} \cite{xavier2021learning} 
used Support Vector Machine (SVM) and \textit{k}-nearest neighbor 
by approximating active transmission constraints and fixing some
reliable variables. 
Recently, Ramesh and Li \cite{ramesh2022feasibility} proposed 
an ML-aided reduced MILP for SCUC problems; they measured the Softmax-
based scores of the ML outputs to fix the commitment variables and
restore the feasibility of
the commitment outputs of the ML model in the post-processing of the
ML inference. 
This paper tackles a similar problem for RACs, but
proposes a unified framework that is trained in an end-to-end manner
to jointly predict generator commitments and active transmission
constraints, and estimate the confidence of the commitment predictions. 

In the context of DC-Optimal Power Flow (DC-OPF), deep learning has 
been shown to approximate optimal solutions successfully, given a 
variety of input configurations (e.g., 
\cite{velloso2021combining,pan2020deepopf}).
Optimal solutions of the non-convex AC-OPF have also been successfully
approximated by deep learning on simpler test cases (e.g.,
\cite{pan2019deepopf,zamzam2020learning,donon2020deep,cengil2022learning,park2022self}). 
The combination of deep learning and Lagrangian duality reducing
constraint violations, together with a feasibility restoration process
using load flows, was shown to find high-quality solutions to
real-world AC-OPF test cases
\cite{fioretto2020predicting,chatzos2020high,chatzos2021spatial}. A
DNN architecture exploiting the patterns appearing in real-world
market-clearing solutions was shown to approximate the solution of
large-scale Security-Constrained Economic Dispatch (SCED) problems
effectively \cite{Chen2022_LearningOptimizationProxies}. Feasibility
restoration, which is imperative in this context, can be obtained by
warm-starting optimization solvers with deep-learning approximations,
(e.g., \cite{baker2019learning,chen2020hot,diehl2019warm}) or finding
active constraints in SCUC (e.g., \cite{xavier2021learning}), DC-OPF
(e.g., \cite{deka2019learning}), and AC-OPF (e.g.,
\cite{robson2019learning,hasan2021hybrid}).
    
Deep learning has also been exploited to find better heuristics. Those
approaches include approximating Newton's method for solving AC-OPFs
(e.g., \cite{baker2020learning}) and estimating parameters in the ADMM
algorithm via a reinforcement learning scheme (e.g.,
\cite{zeng2021reinforcement}). These attempts usually have been
framed in more general optimization settings beyond the power-system
domain. Those include reinforcement learning for general integer
programming (e.g., \cite{tang2020reinforcement}), combinatorial
optimization (e.g., \cite{khalil2017learning}), and MIP (e.g.,
\cite{nair2020solving}).
GNN, a DNN architecture for graph-structured data,
has been utilized especially in AC-OPF
\cite{donon2020deep,diehl2019warm,Owerko2020}. 
    
After Softmax-based entropy to measure confidence was first introduced
in \cite{hendrycks2016baseline}, confidence estimation using epistemic 
uncertainty then became one of the active topics in the deep learning
research community. It can be achieved by estimating data density 
probability (e.g.,
\cite{kingma2016improved,germain2015made,park2021deep}) or uncertainty
quantification (e.g.,
\cite{liang2017enhancing,gal2016uncertainty,gal2016dropout,park2021interpreting}).
Unfortunately, epistemic uncertainty-based confidence estimation on the ML-based output has not been thoroughly studied in power system applications. 

\section{RAC Formulation and Solution Techniques}
\label{sec:preliminaries}

This paper focuses on the RAC problem, a variant of SCUC defined and 
operated by MISO. The detailed formulation is fully stated in \cite{BPM_C}. 
The problem can be modeled as an MILP problem of the form

\begin{subequations}
\label{eq:RAC}
\begin{align}
        (\text{RAC}) \quad \min_{\x, \y} \quad
        & c^{T}\x + q^{T} \y \label{eq:RAC:objective}\\
        \text{s.t.} \quad
        & \x \in \mathcal{X}, \label{eq:RAC:gen:commitment}\\
        & A \x + B \y \geq b, \label{eq:RAC:gen:mixed}\\
        & H \y \geq h, \label{eq:RAC:system}
\end{align}
\end{subequations}

\noindent
where $\x$ denotes the binary commitment/startup/shutdown variables,
and $\y$ denotes the continuous variables such as energy and reserve
dispatches, branch flows, etc. Constraints~\eqref{eq:RAC:gen:commitment} 
denote generator-level commitment-related constraints, such as minimum run/down 
times and \emph{must-run} commitments, and binary requirements on $\x$
variables. Constraints~\eqref{eq:RAC:gen:mixed} denote
generator-level dispatch-related constraints such as minimum/maximum
limits and ramping constraints. Finally, constraints~\eqref{eq:RAC:system} 
denote system-wide reliability constraints,
namely, power balance, minimum reserve requirements, and transmission
constraints. Note that, in MISO's formulation, the
reliability constraints~\eqref{eq:RAC:system} are treated as
\emph{soft}: they can be violated, but doing so incurs a (large)
penalty. 

Indeed, ML-based methodologies for RAC acceleration can be directly applied to SCUC with minor adjustments. However, our attention remains solely on RAC. While RAC and SCUC share similarities in their formulations, RAC requires greater acceleration due to the need to effectively manage significant variations in forecasts in real-time.
    
Despite the significant progress in MILP solver technology and in
constructing tight MILP formulations for SCUC (see, e.g.,
\cite{knueven2020mixed}), industry-size instances remain challenging
\cite{chen2022security}. Transmission constraints are typically handled
in a lazy fashion. Accordingly, the \rac{}
framework uses the iterative optimization procedure suggested in
\cite{xavier2019transmission} as the baseline optimization
algorithm. The algorithm first solves the relaxed problem with no
transmission constraint and then identifies the violations of these
constraints. The $k$ most violated transmission constraints are added
into the relaxed problem, and the process is repeated until there is
no constraint violation. Note that identifying the active set of
transmission constraints beforehand may allow to solve the problem in
a single iteration ideally. The \rac{} framework includes ML techniques to predict this active
set and prior work indicates that such prediction can improve run times
\cite{xavier2021learning}.

\section{The \rac{} Framework}
\label{sec:overview}

The \rac{} framework can be summarized as follows:
\begin{enumerate}
\item Use the GNN-based architecture to predict the generator commitments 
      and active transmission constraints;
\item Use epistemic uncertainty to obtain a confidence value for each 
      commitment prediction, and fix a subset of high confident 
      commitment estimates;
\item Transform a RAC instance with fixed high-confidence commitment 
      variables into the feasible instance using the feasibility 
      restoration step, and;
\item Fix the feasible commitment predictions and seed the active
      transmission constraints in RAC optimization.
\end{enumerate}
The next sections describe these steps in detail. 

\section{The GNN-based Architectures}
\label{sec:architecture}

\begin{figure*}[t!]
\centering
\includegraphics[width=.75\textwidth]{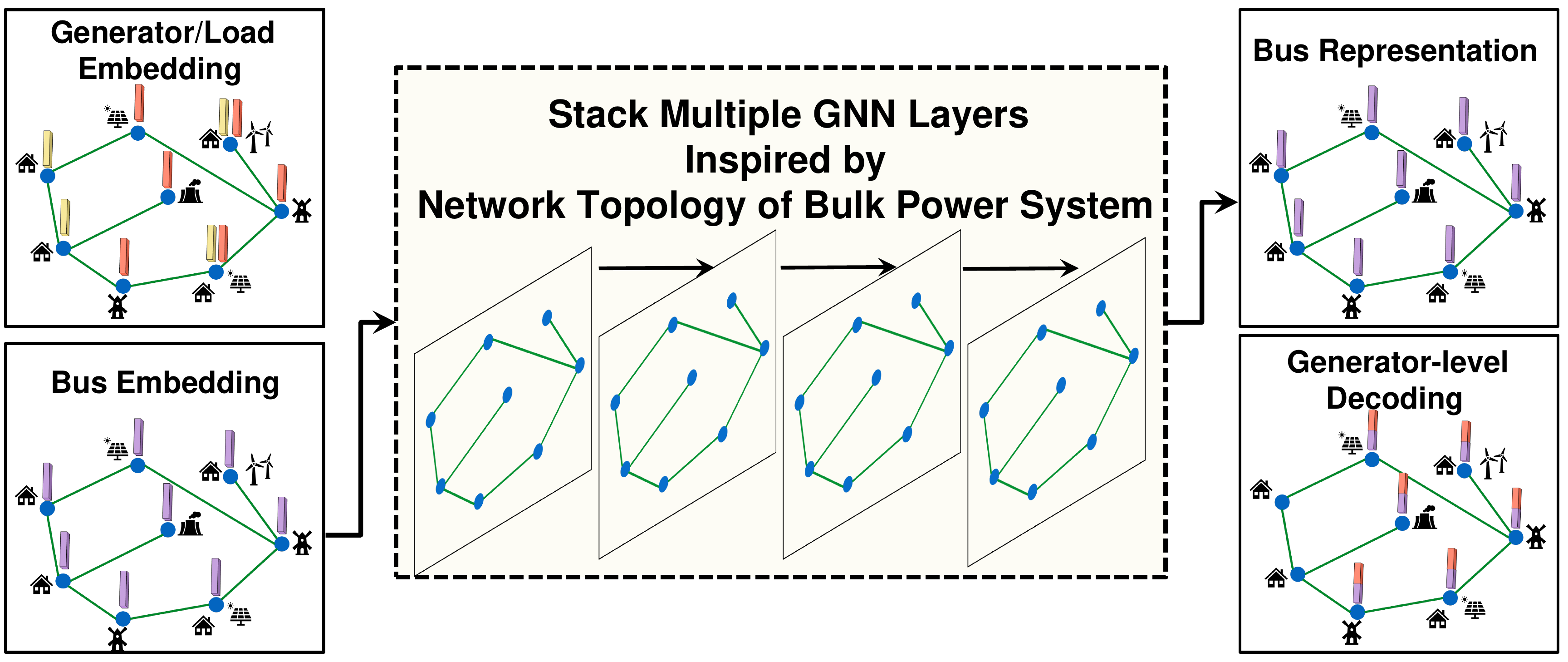}
\caption{An Overview of the Proposed Architecture with GNN Layers. Encoders encode the input configuration vectors for generators and load units to the latent embedded vectors, then these latent embedded vectors are aggregated into the bus-level features (bus embedding). The GNN propagates the information through neighbors and produces the bus representations. Finally, the commitment decoder outputs the optimal commitment prediction for each bus.}
\label{img:gnn}
\end{figure*}

The first component of the \rac{} framework is a novel GNN-based
encoder-decoder architecture depicted in Figure~\ref{img:gnn}. 
Let $\mathcal{B}$ be the
set of bus indices, $\mathcal{G}$ the set of generator indices,
$\mathcal{L}$ the set of transmission line indices, $\mathcal{D}$ the
set of load demand indices, and $\mathcal{T}$ the set of time periods
(commitment intervals). 
The ML model approximating the RAC aims at predicting commitment decisions and active
transmission constraints from the problem inputs.
The problem inputs include the generator and load
configurations $\mathbf{g}_g$ and $\mathbf{l}_l$, detailed in 
Table~\ref{tab:input_config}.
It is assumed that the network topology is fixed and the difference 
between the instances can be described by $\mathbf{g}_g$ and
$\mathbf{l}_l$, which include the forecasting of load demand and 
renewable generation.

The ML model can be viewed as a parametric function
$F_{\theta}$, with trainable parameters $\theta$, that
estimates the binary commitment decisions $\tilde{\mathbf{x}}_g$ 
and the active transmission constraints $\tilde{\mathbf{a}}_l$ as follows:
\begin{equation}
    \left\{ \tilde{\mathbf{x}}_g \right\}_{g\in\mathcal{G}},
    \left\{ \tilde{\mathbf{a}}_l\right\}_{l\in\mathcal{L}}
    = F_\theta\left(\left\{  \mathbf{g}_g\right\}_{g\in\mathcal{G}}, \left\{  \mathbf{l}_l\right\}_{l\in\mathcal{L}}\right).
\end{equation}
\noindent
To account for the high-dimensional and graph-structured nature of
the input data 
$\left\{  \mathbf{g}_g\right\}_{g\in\mathcal{G}}$, $\left\{  \mathbf{l}_l\right\}_{l\in\mathcal{L}}$, 
this paper proposes a GNN-based encoder-decoder architecture. The proposed 
architecture consists of three parts:
\begin{enumerate}

\item The \emph{encoders} encode the input configurations of
      generators and loads into the latent vectors, which are then
      aggregated at the bus level.
\item The core \emph{GNN} layers evolve the embedded vectors at the             buses, and eventually, generate bus representations.
\item The \emph{decoders} outputs the optimal commitment estimates and          active transmission constraint estimates.
\end{enumerate}
\noindent
The rest of this section discusses this architecture in detail.

\subsection{Encoders}
\label{ssec:encoders}

\begin{table}[t!]
    \centering
    \resizebox{0.95\columnwidth}{!}{
    \begin{tabular}{cccc}
    \toprule
    Component                   & Name (BPM notation) & Type & Dimensions \\
    \midrule
    \multirow{20}{*}{Generator}       & EnergyStepWidth     & continuous & $G\times T\times E$ \\
                                      & EnergyStepPrice     & continuous & $G\times T\times E$ \\
                                      & EcoMinLimit         & continuous & $G\times T$         \\
                                      & EcoMaxLimit         & continuous & $G\times T$         \\
                                      & HotToIntTime        & continuous & $G\times 1$         \\
                                      & HotToColdTime       & continuous & $G\times 1$         \\
                                      & InitialOnStatus     & binary     & $G\times 1$         \\
                                      & InitialStatus       & continuous & $G\times 1$         \\
                                      & InputRampRate       & continuous & $G\times T$         \\
                                      & LTSCUCSDRampRate    & continuous & $G\times T$         \\
                                      & LTSCUCSURampRate    & continuous & $G\times T$         \\
                                      & MaxOfflineResponse  & continuous & $G\times T$         \\
                                      & MinDownTime         & continuous & $G\times 1$         \\
                                      & MinRunTime          & continuous & $G\times 1$         \\
                                      & MustRunStatus       & binary     & $G\times T$         \\
                                      & NLOfferPrice        & continuous & $G\times T$         \\
                                      & OffLineSupResOfferPrice & continuous & $G\times T$         \\
                                      & RF                  & binary     & $G\times T$         \\
                                      & RegAvailability     & binary     & $G\times T$         \\
                                      & RegResOfferPrice    & continuous & $G\times T$         \\
                                      & SUCOfferPrice       & continuous & $G\times 1$         \\
                                      & SUHOfferPrice       & continuous & $G\times 1$         \\
                                      & SUIOfferPrice       & continuous & $G\times 1$         \\
                                      & ContResOfferPrice   & continuous & $G\times T$         \\
                                     \midrule
    Load                        & LoadForecast        & continuous & $D\times T$         \\
    \bottomrule
    \end{tabular}
    }
\caption{Input Configurations for DA-FRAC and LAC. $G$ and $D$ are the
number of the generators and load units, respectively. $T$ is the
number of time steps, i.e., $24$ for the DA-FRAC and $12$ for the
LAC.}
\label{tab:input_config}
\end{table}
\paragraph{Generator Encoder}
For each generator, the input feature $\mathbf{g}$
changes from instance to instance.
This input feature $\mathbf{g}$ includes the maximum 
capacity, pricing curve parameters, and so on. 
This input feature $\mathbf{g}$ is 
flattened to be a one-dimensional vector and embedded into a 
higher-dimensional latent vector $\bar{\mathbf{g}}$ through the
\emph{generator encoder} $\text{Enc}_\text{g}$, i.e.,
$
\bar{\mathbf{g}} = \text{Enc}_{\text{g}}(\mathbf{g}).
$

\paragraph{Load Encoder}
Similarly, an input feature for each load unit $\mathbf{l}$ 
(i.e., the load forecast) is embedded into a latent vector
$\bar{\mathbf{l}}$ using the
\emph{load encoder} $\text{Enc}_\text{l}$, i.e.,
$
\bar{\mathbf{l}} = \text{Enc}_{\text{l}}(\mathbf{l}).
$

The generator and load encoders are Multilayer Perceptron (MLP)-based
structures, comprised of fully-connected layers followed by ReLU activations. 
The generator and load encoder structures are formulated as
\begin{equation*}
\begin{aligned}
    \mathbf{h}^{(0)} &= \mathbf{x},\\
    \mathbf{\tilde{h}}^{(l)} &= \mathbf{h}^{(l-1)}\mathbf{W}^{(l)}+\mathbf{b}^{(l)},\\
    \mathbf{h}^{(l)} &= \max\{\mathbf{\tilde{h}}^{(l)}, 0\},
\end{aligned}
\end{equation*}
where $\mathbf{x}$ denotes the generator or load input features.
Also, at each layer $l$, $\mathbf{W}^{(l)}$ is a (trainable) weight
matrix, and $\mathbf{b}^{(l)}$ is a (trainable) bias vector.  
Note that the trainable parameters involved in the generator and load 
encoders are shared across all the generators and load units, 
yielding an efficient structure.

\paragraph{Generating Bus Feature}
Vectors $\bar{\mathbf{g}}$ and $\bar{\mathbf{l}}$ are then
$l_2$-normalized, and the \emph{bus feature} $\mathbf{f}_b\in\mathbb{R}^d$ at
bus $b$ is defined by concatenating the sums of the generator and
load features into
\begin{equation}\label{eq:bus_feature}
\mathbf{f}_b = \left[ \sum_{g\in\mathcal{G}_b}\bar{\mathbf{g}}_g \middle\| \sum_{d\in\mathcal{D}_b}\bar{\mathbf{l}}_d \right],
\end{equation}
where $\left[\cdot \middle\| \cdot\right]$ denotes vector
concatenation, and $\mathcal{G}_b$ and $\mathcal{D}_b$ are the sets of
the generator and load indices attached to bus $b$, respectively. 

\subsection{Graph Neural Networks}
\label{ssec:gnn}
Once the bus features for every bus are prepared, the whole bus 
features are denoted as a matrix form of $\mathbf{X}\in\mathbb{R}^{B 
\times d}$ where each row corresponds to the bus feature $\mathbf{f}$ 
and $B$ represents the number of buses in the system.

In this paper, to embed the output bus features $\mathbf{Y}$ from the input bus feature $\mathbf{X}$, two types of GNN layer are explored: GCN \cite{kipf2016semi} and SIGN \cite{rossi2020sign}. 

\paragraph{Graph Convolutional Network (GCN)}
The power system is viewed as an unweighted undirected graph
$\mathcal{N}=(\mathcal{B}, \mathcal{L})$. The symmetric adjacency
matrix of this graph is denoted by $\mathbf{A}\in\mathbb{R}^{B\times B}$, i.e.,
$A_{ij}=A_{ji}=1$ if there is a transmission line between bus $i$ and
bus $j$. The self-loop added adjacency matrix $\bar{\mathbf{A}}$ is
defined as $\bar{\mathbf{A}}=\mathbf{A}+\mathbf{I}$.

A GCN layer typically utilizes the normalized adjacency matrix
$\tilde{\mathbf{A}}$ defined as:
$\tilde{\mathbf{A}}=\mathbf{D}^{-1/2}\bar{\mathbf{A}}\mathbf{D}^{-1/2}$,
where $\mathbf{D}$ is the degree matrix, i.e.,
$\mathbf{D}=\text{diag}(\sum_{j=1}^B \bar{A}_{1j},\dots,\sum_{j=1}^B
\bar{A}_{Bj})$. The GCN propagates the information to
the neighbor nodes along the edge using graph diffusion operations.
Given the matrix form of the input bus features
$\mathbf{X}\in\mathbb{R}^{B \times d}$, the $d'$-dimensional output bus features
$\mathbf{Y}\in\mathbb{R}^{B\times d'}$ is obtained as
\begin{equation*}
\mathbf{Y} = \text{GCN}(\mathbf{X}|\tilde{\mathbf{A}})= \sigma\left( \tilde{\mathbf{A}}\mathbf{X}\bm{\Theta} \right),
\end{equation*}
where $\sigma(\cdot)$ represents the ReLU activation, and $\bm{\Theta}\in\mathbb{R}^{d\times d'}$ is a matrix of
trainable parameters.
        
The GCN layer only propagates messages to neighboring nodes, i.e.,
1-hop neighbors. To propagate the message throughout the network, a
multiple GCN layers stacked structure can be utilized.  
Using bus embedding
$\mathbf{H}^{(l-1)} \in \mathbb{R}^{B\times d^{(l-1)}}$ as the input
of $l^{th}$ GCN layer, the output is given by
\[
\mathbf{H}^{(l)} = \text{GCN}_l(\mathbf{H}^{(l-1)}|\tilde{\mathbf{A}})= \sigma\left( \tilde{\mathbf{A}}\mathbf{H}^{(l-1)}\bm{\Theta}^{(l)} \right),
\]
where $\bm{\Theta}^{(l)} \in \mathbb{R}^{d^{(l-1)}\times d^{(l)}}$ is
trainable parameters of the layer $l$ and $\mathbf{H}^{(0)} =
\mathbf{X}$. Then, the output of the final GCN layer
$\mathbf{H}^{(L)}$ of the $L$-stacked GCN structure is regarded as the
output bus feature $\mathbf{Y}$. It is still an open question whether
a multiple-stacked GNN structure gives an enhanced performance in
general \cite{rossi2020sign}. Thus, recent research has utilized
different types of diffusion operators to replace the normalized
adjacency matrix and allow messages to propagate beyond 1-hop
neighbors \cite{wu2019simplifying,klicpera2019diffusion}.  For
instance, S-GCN \cite{wu2019simplifying} employs the $p^{th}$ power of
the normalized adjacency matrix to encompass $p$-hop neighbors in its
propagation, i.e., S-GCN is defined as
\begin{equation*}
\mathbf{Y} = \sigma\left( \tilde{\mathbf{A}}^p \mathbf{X}\bm{\Theta} \right).
\end{equation*}

\paragraph{Scalable Inception Graph Neural Network (SIGN)}

Another type of GNN layer, SIGN \cite{rossi2020sign}, uses
Inception-like architecture \cite{szegedy2015going} on
graph-structured data, and generalizes the use of various diffusion
operators in the GCN structure. In particular, SIGN concatenates
multiple output bus features from various diffusion operators. In
addition to the use of the adjacency matrix that only covers the
spatial domain, the SIGN architecture considered in this paper uses
the personalized PageRank (PPR) based spectral diffusion operator
$\mathbf{S}$ \cite{klicpera2019diffusion} and its powers. The SIGN
layer can be formalized as
\begin{equation*}
\mathbf{Y} = \sigma\left(\text{MLP}\left(\xi\left( \bigparallel_{i=1}^p \text{GCN}(\mathbf{X}|\tilde{\mathbf{A}}^i),
                                          \bigparallel_{i=1}^s \text{GCN}(\mathbf{X}|\mathbf{S}^i)                    
                                 \right)\right)\right),
\end{equation*}
where $\xi$ and $\sigma$ are nonlinear activation functions. In
practice, PReLU and ReLU have been used for $\xi$ and $\sigma$,
respectively. Also, MLP denotes fully connected layers followed by
ReLU activations. $p$ and $s$ are hyperparameters representing the
maximum powers used to construct representational features in parallel
for the normalized adjacency matrix and the PPR diffusion operator,
respectively. 
The details on the PPR diffusion operator are provided
in Appendix~\ref{appdx:diffusion_operator}.

\subsection{Decoders}
\label{ssec:decoder}

Denote the output bus feature of bus $j$ from the GNN as
$\bar{\mathbf{f}_j}$, which corresponds to the $j^{th}$ row of
$\mathbf{Y}$. It remains to estimate the optimal commitment decisions and active transmission constraints. 

\paragraph{Commitment Decoder}
Since multiple generators can be located at the
same bus, another generator encoder (independent and distinct from the
previous generator encoder) is used to generate another embedded
vector $\dot{\mathbf{g}}$. This vector is concatenated to the bus
features $\bar{\mathbf{f}_j}$ (with $i\in\mathcal{G}_j$) to obtain a
generator-level representation $\tilde{\mathbf{g}_i}$, i.e.,
$\tilde{\mathbf{g}}_i = \left[ \bar{\mathbf{f}}_j \middle\| \dot{\mathbf{g}}_i \right]$,
Then, the MLP-based \emph{commitment decoder} $\text{Dec}_{\text{g}}$ 
uses $\tilde{\mathbf{g}_i}$ as an input and outputs a
$|\mathcal{T}|$-dimensional vector $\tilde{\mathbf{x}}_i$ that
represents the commitment decision likelihood for the generator
$i$, i.e., $\tilde{\mathbf{x}}_i =
\text{Dec}_{\text{g}}(\tilde{\mathbf{g}}_i)$.  The $t^{th}$ element of
$\tilde{\mathbf{x}}_i$ represents the probability of committing
generator $i$ at time step $t\in\mathcal{T}$, This value is the output
of a sigmoid function attached to the final layer of the decoder
$\text{Dec}_{\text{g}}$.

\paragraph{Transmission Decoder}
The vector $\mathbf{e}_{l}$ of transmission line $l$, 
linking buses $i$ and $j$, can be represented by the
permutation-invariant element-wise summation of the associated output
bus-representations, i.e.,
$
\mathbf{e}_{l} = \bar{\mathbf{f}}_i + \bar{\mathbf{f}}_j.
$
The MLP-based \emph{transmission decoder} $\text{Dec}_{\text{t}}$ uses
vector $\mathbf{e}_{l}$ to estimate whether the
transmission constraints are tight or not, i.e.,
$
\tilde{\mathbf{a}}_{l} = \text{Dec}_{\text{t}}(\mathbf{e}_{l})
$
for all transmissions $l\in \mathcal{L}$.  
Note that $\tilde{\mathbf{a}}$ is a $2|\mathcal{T}|$ dimensional 
vector where $2$ comes from the lower and upper limits of each 
transmission constraint.
Like the commitment decoders, $\tilde{\mathbf{a}}_l$ represents the 
sigmoid-based binary probabilities that the transmission constraints 
of transmission $l$ are active.

\section{Confidence Measurement}
\label{sec:confidence}
The second component of the \rac{} framework is the confidence
measurement of the prediction. Unlike other critical applications such
as semantic segmentation for self-driving AI
\cite{kendall2017uncertainties} or medical image diagnosis
\cite{leibig2017leveraging}, ML approaches in power systems have not
been extensively studied through the lens of uncertainty
quantification. The \rac{} framework remedies this limitation and
provides an epistemic uncertainty-based confidence measure for each
output of the ML model.  
Specifically, to quantify the confidence
in each commitment prediction, \rac{} exploits MC-dropout
\cite{gal2016dropout}, one of the popular tools to measure epistemic
uncertainty (also called model uncertainty) by approximating a
Bayesian neural network \cite{gal2016uncertainty}.  A dropout layer,
at training time, chooses nodes at random, and sets the trainable
parameters associated with those nodes to zero in order to prevent
overfitting during training.  The dropout layer is also used at
inference to express the stochasticity involved in the input data.
That is, instead of providing a single prediction, MC-dropout
generates $T$ of them.
The mean $\mu$ of these $T$ predictions is
then used as the output prediction, while the variance $\sigma$ serves
as a proxy for the epistemic uncertainty, i.e.,
\begin{subequations}
\begin{flalign*}
& \mu(\tilde{x}) = \frac{1}{T}\sum_{t=1}^T \tilde{x}^{(t)}, \\
& \sigma(\tilde{x}) = \frac{1}{T-1}\sqrt{\sum_{t=1}^T \left( \tilde{x}^{(t)} - \mu(\tilde{x}) \right)},\\
\end{flalign*}
\end{subequations}
When the model is to estimate on unseen or unfamiliar input 
instances, $\sigma$ tends to be higher, representing the higher 
epistemic uncertainty involved in the prediction from the model.
Thus, the {\em confidence measure}, which has the reciprocal 
relationship with the epistemic uncertainty, can be defined as
\begin{equation}
\label{eq:confidence_measure}
\text{Conf}(\tilde{x}) := \frac{1}{\sigma(\tilde{x})}.
\end{equation}
\noindent
\rac{} generates confidence values for the commitment of each
generator at each time step. The implementation attaches a dropout
layer with a dropout probability of $0.5$ to the penultimate layer of
the commitment decoder.

\section{The \pro{} Heuristic}
\label{sec:optimization_speedup}

The last component of the \rac{} framework is the feasibility
restoration. It is motivated by a key challenge faced in solving RAC
(or SCUC in general): to find high-quality feasible solutions quickly
\cite{chen2016improving,chen2022security}. \rac{} addresses this
challenge with \pro{} heuristic involving the feasibility restoration 
step, which is guaranteed to return a feasible solution. The \pro{} 
heuristic consists of four components; 1) estimating the active 
transmission constraints and add at the beginning of the iterative 
optimization procedure, 2) fixing generator commitments with
high confidence, 3) executing the feasibility restoration step to repair 
the fixed commitments, and 4) solving the reduced RAC problem w.r.t. the 
remaining continuous variables.

\subsection{Estimating Active Transmission Constraints}

\begin{figure}[t!]
\centering
\includegraphics[width=.65\columnwidth]{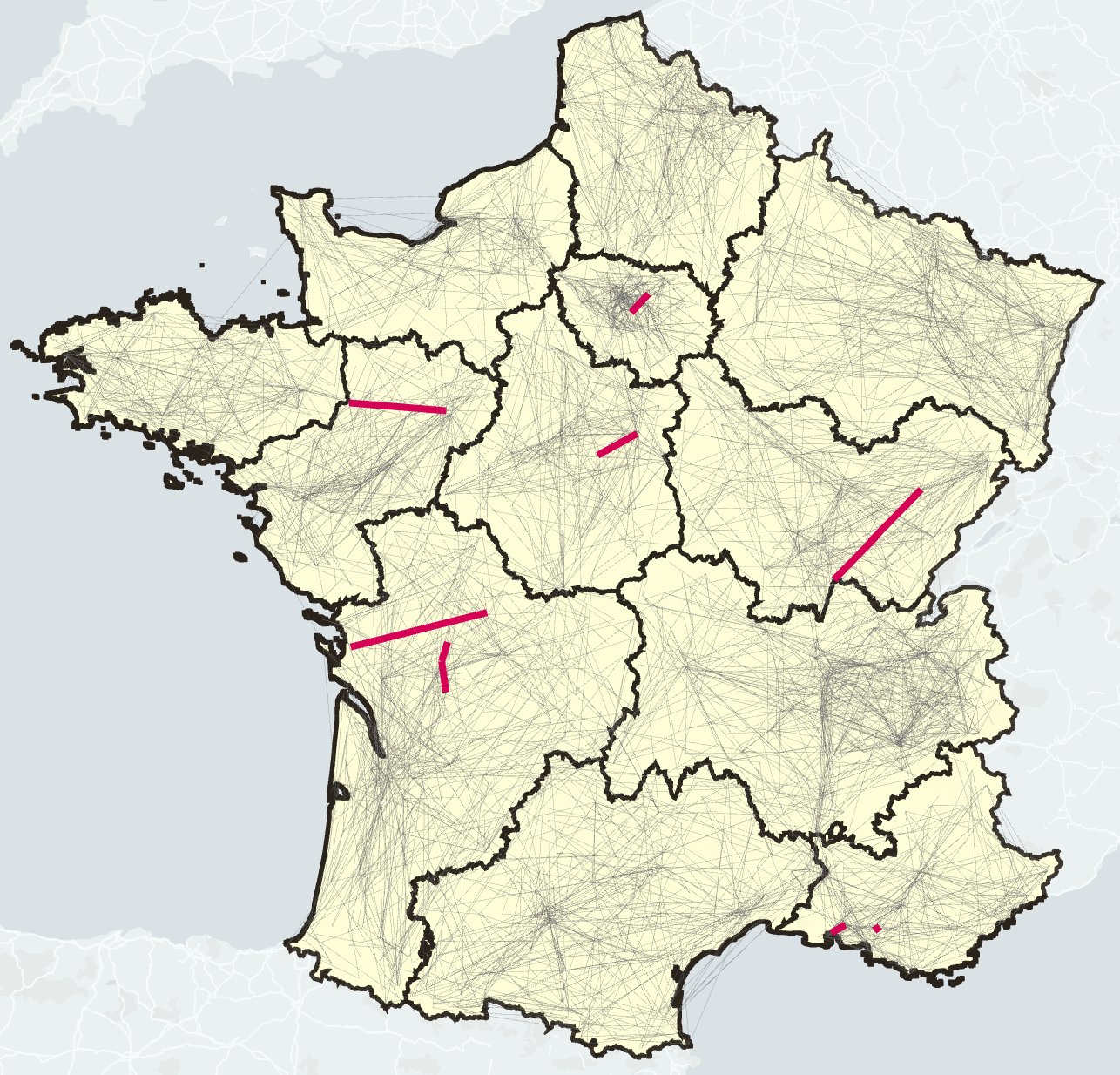}
\caption{Active Transmission Constraints in the RAC Solution. The red
thick lines represent the active constraints, whereas
grey thin lines show non-active transmission constraints.}
\label{img:transmission}
\end{figure}

The first step in the \pro{} heuristic is to estimate the active 
transmission constraints.
Figure~\ref{img:transmission} highlights that few transmission 
constraints are active in an optimal RAC solution. This justifies the 
use of the
iterative optimization procedure \cite{xavier2019transmission} that
first solves the relaxed problem where there is no transmission
constraints and then adds violated transmission constraints lazily
until there is no violation.  To accelerate the optimization
procedure, akin to the approach suggested in
\cite{xavier2021learning}, it is desirable to estimate the active
likely transmission constraints and add those constraints at the
beginning of the iterative procedure in order to decrease the number
of iterations.
Again, the transmission decoder generates a sigmoid-based binary probability as its output to estimate the activeness of a transmission constraint.

\subsection{Fixing Generator Commitments}

The second step of the \pro{} heuristic fixes the binary variables
corresponding to \emph{confident} commitment decisions. The confidence
level needed to fix the commitment decisions is a hyper-parameter
that is studied in Section \ref{sec:experiments}.
    
\subsection{Feasibility Restoration}

Fixing confident commitments may result in an infeasible RAC problem 
instance. Specifically, the prediction $\tilde{\x}$ may conflict with the
commitment-level constraints~\eqref{eq:RAC:gen:commitment} in the RAC
formulation. To address this issue, the \rac{} framework repairs
the prediction $\tilde{\x}$ to obtain commitment decisions $\hat{\x}$
that satisfy \eqref{eq:RAC:gen:commitment}. Then, the commitment
decisions are fixed to $\hat{\x}$, and the RAC is re-optimized,
yielding $\hat{\y}$ such that $(\hat{\x}, \hat{\y})$ is feasible for
Problem \eqref{eq:RAC}.  The rest of this section describes the
repair step and proves that it runs in polynomial times.

Let $\tilde{\x}$ be the optimal commitment estimates to a RAC
instance. This prediction does not guarantee that $\tilde{\x}$ will
satisfy the combinatorial constraints \eqref{eq:RAC:gen:commitment}.
To alleviate this issue, the paper introduces a repair step which,
akin to a projection step, seeks a feasible $\hat{\x} \in \mathcal{X}$
that is close to $\tilde{\x}$. Formally, the repair step solves
\begin{subequations}
\label{eq:repair}
\begin{align}
\hat{\x} \in \arg\min_{\x} \quad
   & \mathcal{H}(\x, \tilde{\x}) \label{eq:repair:objective}\\
   \text{s.t.} \quad
    & \x \in \mathcal{X} \label{eq:repair:gen:commitment},
\end{align}
\end{subequations}

\noindent
where $\mathcal{H}(\x, \tilde{\x})$ is the Hamming distance between
$\x$ and $\tilde{\x}$.

\begin{theorem}
\label{thm:repair:polytime}
Problem \eqref{eq:repair} can be solved in polynomial time.
\end{theorem}
\begin{proof}
First, because commitment constraints are formulated at the generator
level, Problem \eqref{eq:repair} can be decomposed into $G$
single-generator problems, where $G$ is the number of
generators. Second, because $\x, \tilde{\x}$ are binary vectors,
objective \eqref{eq:repair:objective} is linear.  Thus, Problem
\eqref{eq:repair} reduces to optimizing a linear objective over the
commitment constraints for a single generator.  This is solvable in
polynomial time using the dynamic-programming algorithm of Frangioni
and Gentile \cite{FrangioniGentile2006_PolyTimeSingleUnitCommitment}.
\end{proof}
        
\begin{theorem}
\label{thm:PRO:polytime}
The \pro{} heuristic returns a feasible solution $(\hat{\x}, \hat{\y})$ in polynomial time.
\end{theorem}
\begin{proof}
By Theorem \ref{thm:repair:polytime}, a feasible $\hat{\x}$ can be
obtained in polynomial.  Then, recall that constraints
\eqref{eq:RAC:system} are soft.  In addition, the structure of
constraints \eqref{eq:RAC:gen:mixed} is such that, for any $\x$ that
satisfies \eqref{eq:RAC:gen:commitment}, there exists a vector $\y$
that satisfies \eqref{eq:RAC:gen:mixed}. For instance, it suffices
to set the reserve dispatch of each generator to zero, and its energy
dispatch to its minimum output at every time period.  Finally, once
the binary variables $\x$ are fixed, Problem \eqref{eq:RAC} reduces
to a linear program, which can be solved in polynomial time to obtain $\hat{\y}$.
\end{proof}
        
\section{Experimental Results}
\label{sec:experiments}

\subsection{Experimental Settings}
\label{ssec:experiment_setting}

\subsubsection{The Case Studies}
Experiments were conducted on the RTE system in France. The system
contains 8965 transmission lines and 6708 buses where 1890 generators
and 6262 load units are located.  Solar/wind power generation and load
demand forecast were obtained through LSTM-based models
\cite{hochreiter1997long} trained on the publicly-available ground
truth obtained from RTE \cite{eco2mix}.  The stochastic forecast
scenarios are generated via MC-dropout \cite{gal2016dropout}.  Supply
offer bids were synthetically generated by the approach proposed in
\cite{Chatzos2022_TimeSeriesReconstruction}.
                    
\begin{figure}[t!]
\centering
 \includegraphics[width=.80\columnwidth]{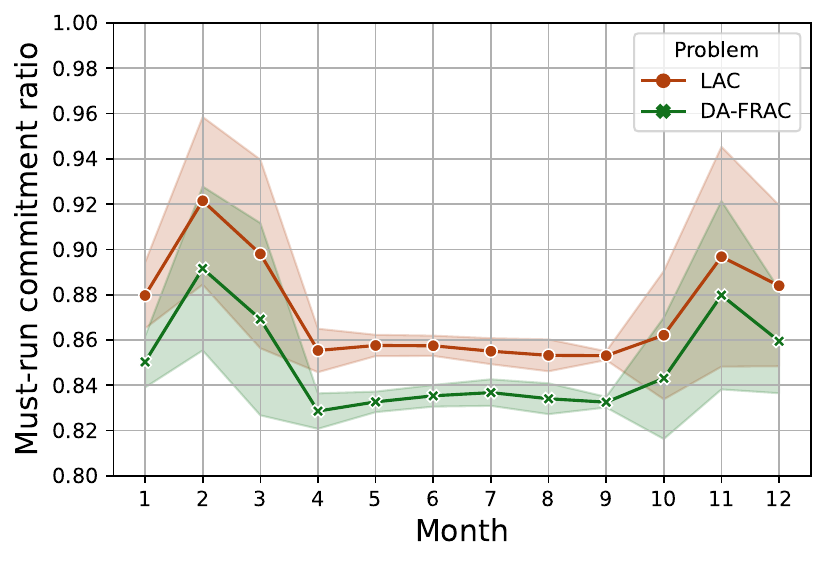}
\caption{Ratio of the Must-run Commitments ($x_{gh}^{\text{prev}}=1$) in DA-FRAC and LAC Instances. The shaded region represents the standard deviation.}
\label{img:must_run}
\end{figure}
                
Two distinct types of RAC problems are studied: DA-FRAC and LAC.  For
gathering DA-FRAC instances, the DA market (DA-SCUC) is first executed.  
The must-run commitments for the DA-FRAC are then extracted from the
DA-SCUC solution.  For gathering LAC instances, the DA-FRAC is first
executed and the must-run commitments are extracted from its solution.
A LAC instance is solved every 15 minutes and it uses all the
commitment decisions of the DA-FRAC and prior LAC executions as the
must-run commitments.  Figure~\ref{img:must_run} shows that the majority
of the commitment decisions are already determined by prior
optimization decisions.  Ratios of the must-run commitments are above
80\% and 85\% for DA-FRAC and LAC, respectively.
    
\subsubsection{Data Generation}

The formulations of the DA-FRAC, LAC, and DA-SCUC were modeled with
JuMP \cite{JuMP} and solved through Gurobi 9.5 \cite{gurobi}.  To
gather the instances, the optimizations were solved on Linux machines
in the PACE Phoenix cluster at Georgia Tech \cite{PACE}.  The data
instances are gathered by setting the optimality gap to $0.1\%$ and
using 2 threads and $8GB$ RAM.  The time limit was set to 3600
seconds and the parameter \texttt{MIPFocus} was set to 1 for focusing
on finding feasible solutions.  The remaining parameters use the
default values.  Only the instances with the optimal solutions that
are obtained within the time limit are included in the dataset for
stable training.  When generating instances, transmission
constraints are added in an iterative manner as
suggested in \cite{xavier2019transmission}.  The solution process
starts with no transmission constraints.  At each iteration, the 15
most violated transmission constraints are added into the formulation
sequentially until no violations are observed.
                    
The number of data instances for each month is around 9000 for the
DA-FRAC and 8600 for the LAC.  The training and testing procedures are
performed on a monthly basis, i.e., the training is performed on data
instances of a month, and the testing is conducted on the data
instances of the following month.  Thus, in 2018, 11 independent
models for each month (February to December) are trained and tested.  
Note that this experimental setting inherently features a 
\emph{distribution shift} \cite{quinonero2008dataset}, since the data distributions of the training and testing datasets are different.  
It mimics the natural way of deploying the models in real-world settings, 
using historical data to infer the upcoming commitment decisions.  For 
each month of training, 100 instances are extracted and used for 
validation.
1000 testing instances are also sampled from the next month's instances for measuring and reporting the performance results.
            
\subsubsection{Training Configuration}
The proposed GNN-based architecture and other neural network baselines
were implemented and trained using PyTorch.  
The proposed architecture is trained in an end-to-end manner by 
minimizing binary cross-entropy loss with respect to the commitment 
estimates and active transmission constraint estimates in a supervised 
way. The minibatch size was set to 8 for training. The AdamW optimizer
\cite{loshchilov2017decoupled} with a learning rate of
$1\mathrm{e}{-4}$ and a weight decay of $1\mathrm{e}{-6}$ was used for
updating the trainable parameters.  Early stopping was applied with a
the patience of 5, while the learning rate decayed by $0.95$ when the
validation loss was worse than the best validation loss at every
epoch.  The maximum epoch was set to 20.  The training and testing
process was performed using Tesla V100 GPU on machines with Intel CPU
cores at 2.7GHz.
The architectural design of the proposed GNN-based
encoder-decoder structure is detailed in
Appendix~\ref{appdx:arch_detail}.
Other details on the training
process are elaborated in Appendix~\ref{appdx:training}.

\subsection{Commitment Prediction Accuracy}
\label{ssec:experiment_result1}
                
\begin{figure*}[!t]
    \centering
    \includegraphics[width=0.49\textwidth]{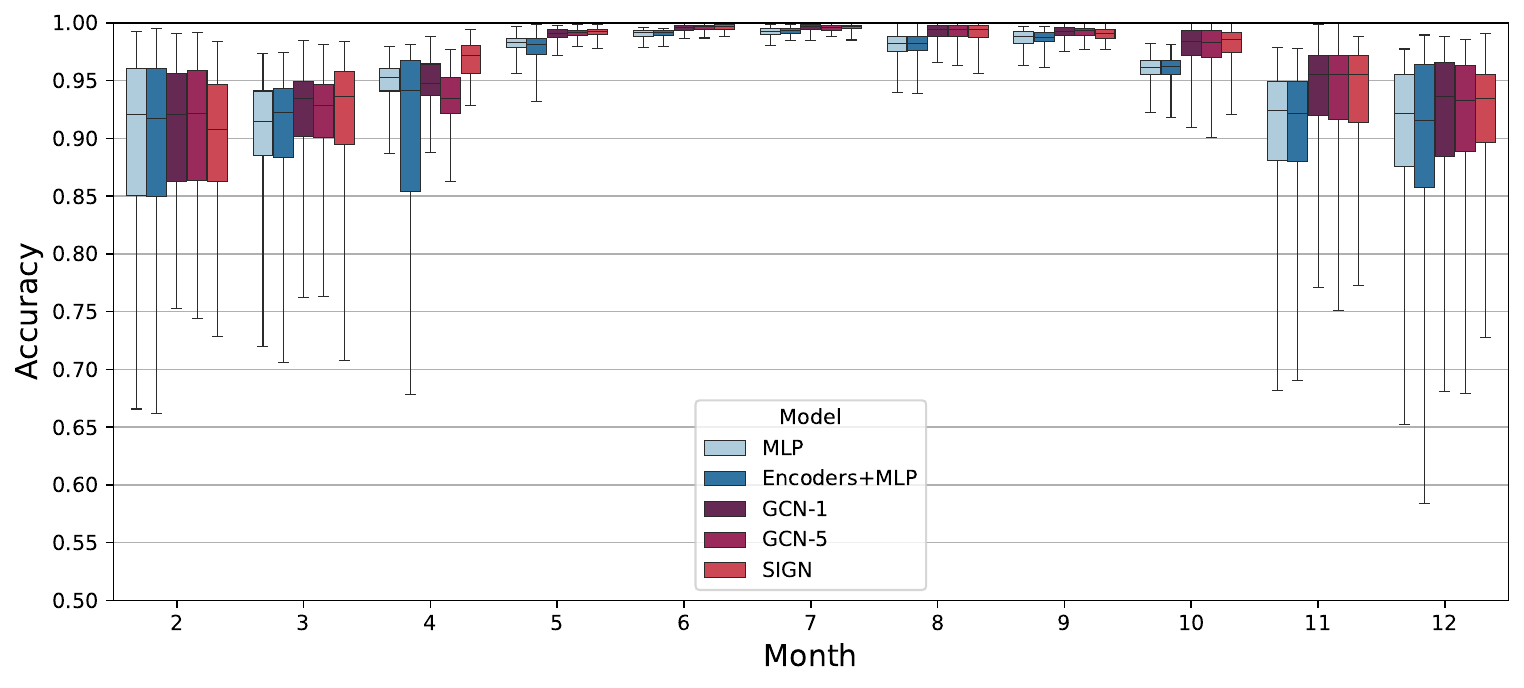}
    \includegraphics[width=0.49\textwidth]{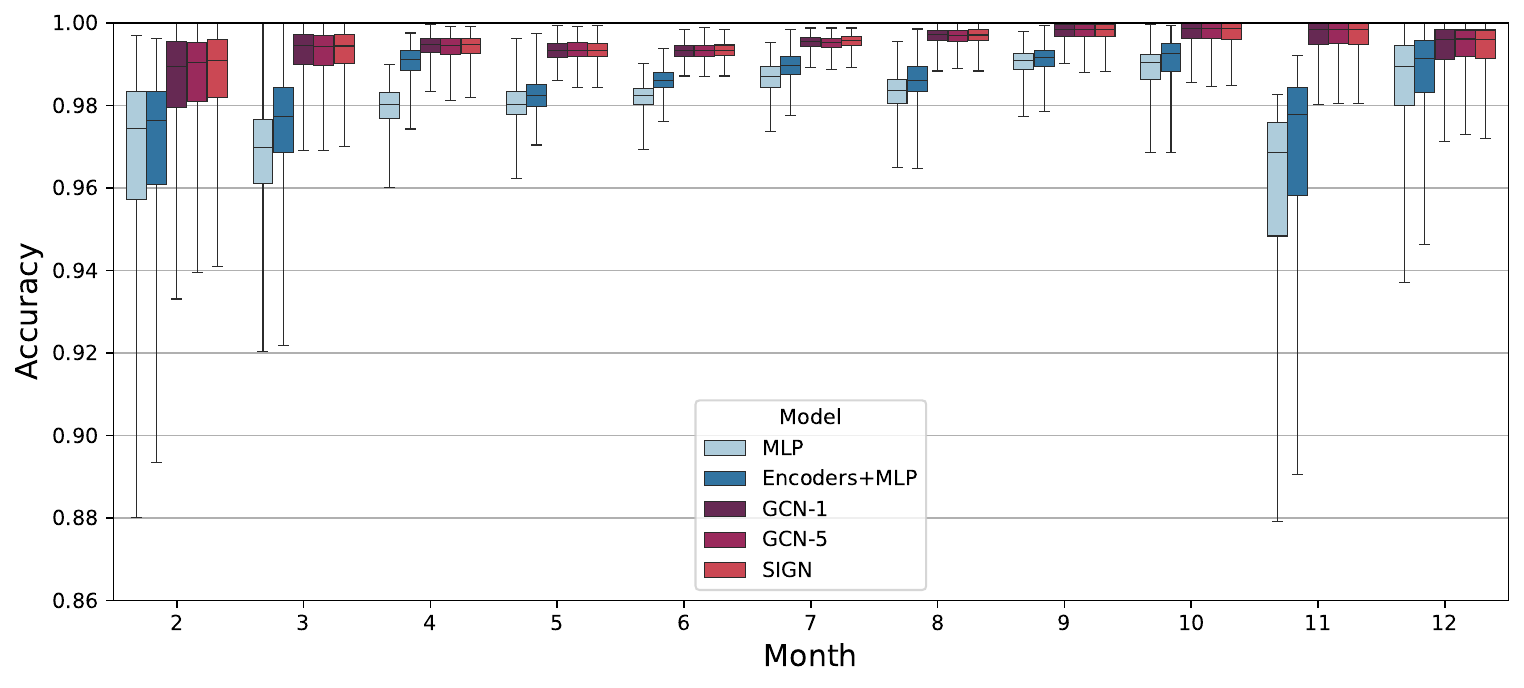}
    \caption{Monthly Performance for DA-FRAC (Left) and LAC (Right).}
    \label{img:month_performance}
\end{figure*}
                  
This section reports the performance of the optimal commitment
predictions of the proposed GNN-based models as well as various
baselines.
The \rac{} framework was tested with several GNN layers: While
the encoder and decoder structures remain the same for every
architecture, the GNN layers to propagate bus representations are
different: they are \texttt{SIGN}, \texttt{GCN-1} (a single GCN layer), and \texttt{GCN-5} (5 stacked GCN layers).
\subsubsection{Baselines}

The \rac{} architectures are compared to two neural network baselines:
\texttt{MLP} has three fully-connected layers of 512 hidden nodes and
flattens all the input features; \texttt{Encoders+MLP} uses the
encoder structures of \rac{}, flattens the embedded generator/load
features, and applies 3 fully-connected layers, followed by 
ReLU activations, with 512 hidden nodes instead of using the GNN 
and decoder structures of \rac{}. 
                
The \rac{} framework is also compared to the following shallow models
taken from scikit-learn \cite{scikit-learn}. \texttt{SVM-poly} is an
SVM model with a polynomial kernel, \texttt{SVM-rbf} is an SVM model
with radial basis function-based kernel, and \texttt{RF} is a
random-forest model.  All those shallow models use the default
parameter settings. Those shallow models are not scalable to the case
study and hence a \naive{} autoencoder (AE) was used to decrease the
dimensionality of the input feature space. AE has three MLP layers for
each encoder and decoder. The encoder embedding vector has a
dimensionality of 0.2\% of that of the input features. The number of
hidden nodes is set to the mean of dimensions of the input and
embedding features. This AE model is also trained using the AdamW
optimizer with a learning rate of $5\mathrm{e}{-5}$ up to 20 epochs
with a minibatch of 16 instances. When training, must-run commitments
are masked out to focus on the non-must-run commitments.
            
\subsubsection{Accuracy Results}
    
\begin{table}[t!]
\centering
        \begin{tabular}{l|ccc}
            \toprule
            Model        & Accuracy     & AUROC \\
            \midrule
            \texttt{SVM-poly}     & 0.895(0.692,0.951) & 0.853(0.650,0.960) \\
            \texttt{SVM-rbf}      & 0.898(0.674,0.949) & 0.855(0.651,0.961) \\
            \texttt{RF}           & 0.896(0.702,0.949) & 0.860(0.645,0.968) \\
            \midrule
            \texttt{MLP}          & 0.960(0.907,0.992) & 0.870(0.626,0.955) \\
            \texttt{Encoders+MLP} & 0.960(0.905,0.993) & 0.870(0.626,0.955) \\
            \midrule
            \texttt{GCN-1}        & \textbf{0.981}(0.915,0.996) & 0.869(0.748,0.976) \\
            \texttt{GCN-5}        & 0.980(0.915,0.996) & 0.892(0.742,0.983) \\
            \texttt{SIGN}         & \textbf{0.981}(0.914,0.996) & \textbf{0.905}(0.776,0.990) \\
            \bottomrule
        \end{tabular}
        \caption{Overall Accuracy and AUROC Performance Results for
          DA-FRAC. The bounds of the 95\% confidence intervals are
          presented between parenthesis. The best values are in bold.}

        \label{table:performance_frac}
        \end{table}
            
        \begin{table}[t!]
        \centering
        \begin{tabular}{l|ccc}
            \toprule
            Model        & Accuracy     & AUROC \\
            \midrule
            \texttt{SVM-poly}     & 0.913(0.666,0.949) & 0.916(0.640,0.958) \\
            \texttt{SVM-rbf}      & 0.912(0.692,0.951) & 0.883(0.664,0.961) \\
            \texttt{RF}           & 0.916(0.733,0.948) & 0.905(0.683,0.961) \\
            \midrule
            \texttt{MLP}          & 0.982(0.966,0.991) & 0.965(0.678,0.986) \\
            \texttt{Encoders+MLP} & 0.986(0.973,0.991) & 0.980(0.839,0.993) \\
            \midrule
            \texttt{GCN-1}        & \textbf{0.994}(0.990,0.998) & \textbf{0.995}(0.975,1.000) \\
            \texttt{GCN-5}        & \textbf{0.994}(0.990,0.998) & \textbf{0.995}(0.973,0.999) \\
            \texttt{SIGN}         & \textbf{0.994}(0.990,0.998) & \textbf{0.995}(0.976,0.999) \\
            \bottomrule
        \end{tabular}
        \caption{Overall Accuracy and AUROC Performance Results for LAC. The bounds of the 95\% confidence intervals are presented between parenthesis. The best values are in bold.}
        \label{table:performance_lac}
        \end{table}
                
The prediction results for DA-FRAC and LAC are depicted in Tables
\ref{table:performance_frac} and \ref{table:performance_lac}. The
tables report the mean accuracy together with 95\% confidence
intervals. The metrics are calculated using bootstrapping with 1000 
samples from the testing set with replacement. 
The performance measures are the averaged accuracy and
the area under the receiver operating characteristic curve (AUROC). In
all cases, GNN-based models are superior to the swallow and MLP-based
models in terms of accuracy and AUROC.  There is no significant
difference in performance among the GNN-based models. The prediction
accuracy is higher for the LAC than for the DA-FRAC, which is
explained by the smaller variability in forecasts and the shorter time
horizon of the LAC.

Figure \ref{img:month_performance} depicts the monthly performance of
the models (the shallow models are excluded because of their poor
performances).  It highlights that the winter season (from October to
March) is more challenging to predict and has more outliers. The
figure also indicates that the GNN-based architectures outperform their 
MLP-based counterparts by a significant margin.

Initially, it was anticipated that SIGN would exhibit superior performance compared to other GNN-based architectures, as SIGN leverages more comprehensive operators to distill information from the entire power grid's features. However, the results indicate that there are no notable distinctions among GNN-based architectures. This can be attributed to the fact that the power grid can be characterized as a \emph{small-world} network \cite{watts1998collective}, where most nodes can be reached from every node with a limited number of hops. As such, the generation commitment can be estimated with the localized features; the loads and the generator's features from neighboring nodes, rather than from distant nodes.
        
\subsection{Commitment Prediction Confidence}
\label{ssec:experiment_confidence}
    
\begin{figure}[!t]
\centering
\includegraphics[width=0.80\columnwidth]{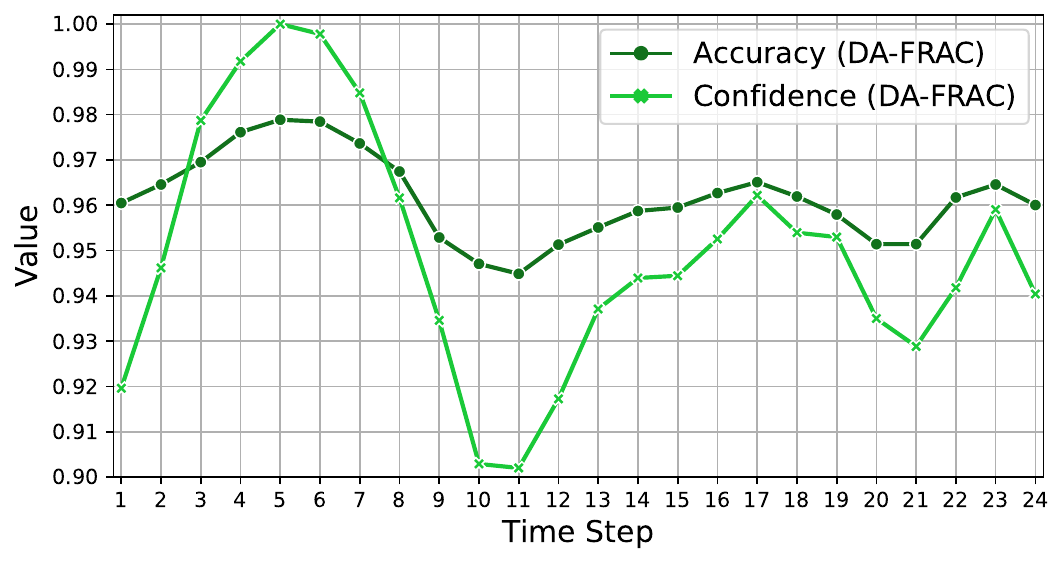}
\vspace*{2mm}
\includegraphics[width=0.80\columnwidth]{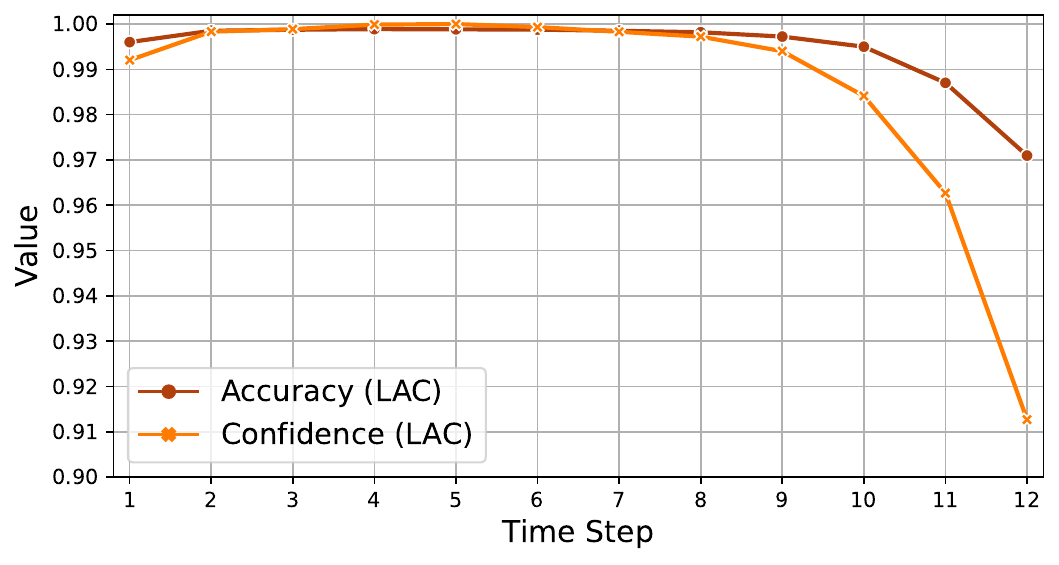}
\caption{Average Testing Accuracy and Associated Confidence Values for
  the Commitment Decisions at Each Time Step of DA-FRAC (Top) and
  the LAC (Bottom). Confidence values are normalized (i.e., divided by
  the maximum confidence value).}
\label{img:confidence_timestamp}
\end{figure}
        
\begin{figure}[!t]
\centering
\includegraphics[width=0.80\columnwidth]{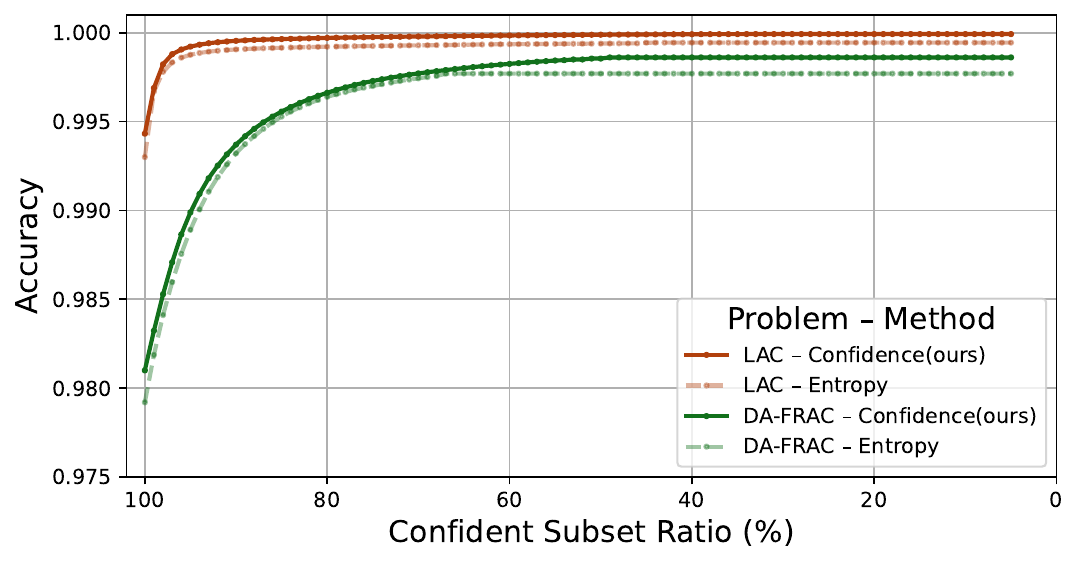}
\caption{Performance on the Confident Subsets and Whole Set. The x-axis shows the ratio of the number of commitments in the confident subset divided by the total number of commitments (denoted in percentage). Confidence used in this work is compared with the entropy measure.}
\label{img:confidence_subset}
\end{figure}

Among the GNN-based models, the \texttt{SIGN} model was chosen for the 
rest of the experiments.  
Figure~\ref{img:confidence_timestamp}
shows the average testing accuracy and its corresponding normalized
confidence values from Eq.~\eqref{eq:confidence_measure} of the 
commitment predictions at each time step for the DA-FRAC and the LAC. 
Recall that
the numbers of time steps for the DA-FRAC and the LAC are 24 and
12. For the LAC, the accuracy of the predictions decreases as the time
step increases. This comes from the fact that the LAC is run over a
rolling horizon during operations. Earlier steps have more must-run
commitments and more accurate forecasts. For the DA-FRAC, each
instance predicts the next day, the forecasts for each hour are more
consistent, and fewer must-run commitments are imposed. In this case,
the accuracy is consistent within the range between $0.95$ and
$0.99$. Interestingly, in both cases, the confidence value is high and
in agreement with the accuracy.  For instance, in LAC, the average
confidence value at step 12 is significantly lower than
other steps, which is indicative of less accurate results. This is
confirmed by the accuracy results. Note that the confidence
value is obtained without relying on the ground truth.

The \rac{} framework uses commitment predictions to fix binary
variables, reducing the dimensionalities of
these optimization problems. Moreover, the confidence values
associated with the predictions provide valuable information about
which predictions must be used in this process in order to balance the
quality of the solutions and the dimensionality reduction. Using lower
confidence predictions may deteriorate the quality of the solutions,
while only using the predictions with the highest confidence may not
provide significant computational benefits. 
Figure~\ref{img:confidence_subset} explores this trade-off. The x-axis
captures the percentage of highly confident commitment predictions 
selected to fix variables in the optimization models, which is called the 
{\em confident subset}. The y-axis captures the average accuracy of the 
selected predictions. The percentages are ranked in decreasing order in 
Figure~\ref{img:confidence_subset}: a percentage of 100\% means that all 
the predictions will be used; a percentage of 30\% means that the 
confident subset consists of the 30\% of predictions with the highest 
confidence values. The figure shows how fast the average accuracy 
increases monotonically as the size of the confident subset decreases for 
both the FRAC and the LAC cases. For the LAC, the accuracy can increase to
around $1.000$ as soon as the confident subset contains less than 50\%
of the predictions. 
Similarly, for the DA-FRAC, the accuracy increases to around $0.998$ if 
low-confident commitment estimates are rejected, whereas when all 
commitment estimates are concerned, the accuracy was $0.981$.
These predictions based on the MC-dropout-based confidence measure are likely to be accurate, leading to an attractive tradeoff between solution quality and
dimensionality reduction in the reduced optimization.

Figure~\ref{img:confidence_subset} also includes the binary entropy (denoted as {\em Entropy}) for comparison. 
This measure is directly induced by the softmax-based probability 
\cite{hendrycks2016baseline}, which has been utilized in the previous ML-based approaches in power system applications \cite{ramesh2022feasibility,xavier2021learning}. 
The figure shows that the proposed confidence measure (based 
on MC-dropout) performs better than the binary entropy to classify the confident subset. This coincides with the comparison results in  \cite{leibig2017leveraging} in the medical domain.

\subsection{Prediction of Active Transmission Constraints}
\label{ssec:experiment_transmission}
    
\begin{table}[!t]
\centering
\begin{tabular}{llccc}
\toprule
Problem & Method & Recall$\uparrow$ & \#Pred Pos.$\downarrow$ & \#Pos. \\
\midrule
\multirow{2}{*}{DA-FRAC} 
        & Baseline & 0.822(0.057)          & 965.24          & \multirow{2}{*}{125.29}\\
        & \rac{}   & \textbf{0.914}(0.063) & \textbf{409.91} \\
\midrule
\multirow{2}{*}{LAC} 
        & Baseline & 0.979(0.008)          & 242.46          & \multirow{2}{*}{70.94}\\
        & \rac{}   & \textbf{0.981}(0.034) & \textbf{141.18} \\
\bottomrule
\end{tabular}\\
 \caption{Recall Performance of Active Transmission Constraint Prediction.
    Column {\em \#Pred Pos.} represents the averaged number of predicted active transmission constraints. Column {\em \#Pos.} denotes the averaged number of actual active transmission constraints.}
 \label{table:results:transmission_constraint}        
 \end{table}

Actual unit-commitment problems have few active transmission
constraints and previous works have shown that they can be predicted
with reasonable accuracy by identifying those transmissions
\cite{roald2019implied,xavier2021learning}. To evaluate the
performance of the \rac{} framework in predicting active
constraints, this paper adopts baselines that exploit prior runs of
the RAC model, i.e., the FRAC of the previous day or the LAC that
was run 15 minutes before. 
In the baseline, all the constraints are considered active if they are included in a transmission line that had at least one active constraint in the prior run.
In \rac{}, transmission constraint predictions with above 0.1 probability are regarded as being active.

Table~\ref{table:results:transmission_constraint} presents
experimental results showing the benefits of predicting active
constraints compared to the baselines. 
Those predictions are expected to obtain high recall values while having a small number of active constraint predictions.
The recall value, i.e., the true positive rate, is
the fraction of correctly predicted active constraints among all
constraints that are actually active.
The recall value is an appropriate measure because the false negatives (the constraints predicted as not active, but active actually) should be added to the formulation subsequently, which leads to excessive time for solving the problem.

On average, there are about
125 active constraints in the DA-FRAC and 71 in the LAC. On average,
the baseline isolates 965.24 constraints (DA-FRAC) and 242.46
constraints (LAC) that can be active for recall values of 0.822
(DA-FRAC) and 0.979 (LAC). In contrast, \rac{} identifies 409.91 (DA-FRAC)
and 141.18 (LAC) possible active constraints on average.
The recall values are 0.914 and 0.981 for the DA-FRAC and LAC, respectively. 
Hence, compared to the baseline, \rac{} provides better
recall values (especially for the DA-FRAC) and adds fewer constraints
to the optimization model, demonstrating its practical value.
More detailed results are included in Appendix~\ref{appdx:active_transmission}.

\subsection{Optimization Speedups}
\label{ssec:speedup}

\begin{table}[!th]
\centering
\begin{tabular}{lcrrrr}
\toprule
&&&&\multicolumn{2}{c}{CPU time}\\
\cmidrule(lr){5-6}
Method & \%Fixed$^{\dagger}$ & Feas(\%) & Gap(\%) & Total(s) & Scaled$^{\ddagger}$ \\ 
\midrule\midrule
\multicolumn{6}{c}{DA-FRAC}\\ 
\midrule\midrule
Baseline  & -- & 100.00 & 0.00 & 126.06 & 1.00 \\ 
\midrule
\multirow{5}{*}{\begin{tabular}{@{}l@{}}\rac \\ w/o feasibility \\ restoration \end{tabular}}
         &  30 & 100.00 & 0.01 &  48.93 & 0.38 \\
         &  50 & 100.00 & 0.02 &  47.56 & 0.37 \\
         &  70 & 100.00 & 0.06 &  45.07 & 0.35 \\
         &  90 &  89.61 & 0.07 &  35.82 & 0.30 \\
         & 100 &  31.17 & 0.12 &  19.19 & 0.25 \\ 
\midrule
\multirow{5}{*}{\rac{}}
         &  30 & 100.00 & 0.01 &  48.40 & 0.38 \\
         &  50 & 100.00 & 0.02 &  47.36 & 0.37 \\
         &  70 & 100.00 & 0.06 &  45.03 & 0.35 \\
         &  90 & 100.00 & 0.16 &  38.81 & 0.30 \\
         & 100 & 100.00 & 0.77 &  31.56 & 0.24 \\
\midrule\midrule
\multicolumn{6}{c}{LAC}\\ 
\midrule\midrule
Baseline   &  -- & 100.00 & 0.00 &  42.71 & 1.00 \\
\midrule
\multirow{5}{*}{\begin{tabular}{@{}l@{}}\rac \\ w/o feasibility \\ restoration \end{tabular}}
         &  30 & 100.00 & 0.00 &  26.44 & 0.62 \\
         &  50 & 100.00 & 0.00 &  25.08 & 0.59 \\
         &  70 &  99.53 & 0.01 &  25.04 & 0.59 \\
         &  90 &  98.59 & 0.01 &  21.87 & 0.51 \\
         & 100 &  84.51 & 0.09 &  21.59 & 0.53 \\
\midrule
\multirow{5}{*}{\rac{}}
         &  30 & 100.00 & 0.00 &  25.45 & 0.60 \\
         &  50 & 100.00 & 0.00 &  24.82 & 0.58 \\
         &  70 & 100.00 & 0.01 &  23.67 & 0.56 \\
         &  90 & 100.00 & 0.01 &  22.73 & 0.53 \\
         & 100 & 100.00 & 0.09 &  20.94 & 0.49 \\
\bottomrule
\end{tabular}\\
$^{\dagger}$percentage of commitments fixed in \rac{}. \\
$^{\ddagger}$relative to baseline. 
CPU Time is computed on the feasible instances.
\caption{Optimization Speedup Performance of the \rac{} Framework. The
baseline is the optimization baseline using \cite{xavier2019transmission};
the \rac{} performances are reported for various confidence levels of the 
commitment estimates fixed (shown in column {\em \%Fixed}). Column {\em 
Feas} represents the fraction of the feasible instances. Column {\em Gap}
represents the geometric mean of the optimization gaps. Column {\em Total}
is the geometric mean of the CPU running times.}
 \label{table:results:optimization}        
 \end{table}
    
The results for the optimization speedups are reported in
Table~\ref{table:results:optimization}.  The baseline is the iterative
optimization procedure \cite{xavier2019transmission} with $k=15$
(adding 15 transmission constraints at each iteration of the overall
procedure) as suggested in \cite{xavier2021learning}. According to
\cite{barry2022risk}, at MISO, the DA-FRAC and LAC have to be solved
within 40 and 10 minutes. The results in elapsed times of the baseline
for DA-FRAC and LAC are $126.06$ and $42.71$ seconds. These times
are higher than those obtained for the MISO system because the RTE
system is significantly larger and more congested in general. For
\rac{}, Table ~\ref{table:results:optimization} reports the results
when the confident subset contains 0\%, 30\%, 50\%, 70\%, 90\%, and
100\% of the predictions. For the LAC, the results show that enforcing
the predictions (even all of them) incurs only a small loss in
solution quality, e.g., 0.01\% for 90\% of the predictions and 0.09\%
for 100\%.  Moreover, the computing times are reduced by about
50\%. For the DA-FRAC, committing up to 90\% of the predictions
results in a small quality loss and reduces the execution times by
more than a factor of 3. It reduces the execution times by a factor of
more than 4, if a quality loss of $0.77\%$ is acceptable.

If the feasibility restoration step is not included in the \rac{}
framework, fixing the commitment predictions could produce
infeasible solutions. In fact, when all commitment variables are
fixed to their predictions, only 31.17\% of instances are feasible
for the DA-FRAC cases. The feasibility restoration step, which takes
polynomial time, transforms infeasible solutions into feasible
solutions with a slightly worse optimality gap for the
DA-FRAC. For the LAC, the feasibility restoration does not
affect the optimality gap significantly.

It is important to emphasize that the optimization speedup results
rely on the three components of \rac{}: the commitments of generators
for which the predictions are highly confident, the predictions of the
active transmission constraints, and the feasibility restoration.
More detailed results on the
quality/performance tradeoffs and the role of each component are given
in Appendix~\ref{appdx:optimization_speedup}.

\section{Conclusion}
\label{sec:conclusion}

This paper proposed the \rac{} framework to speed up the solving of
RAC problems and, in particular, the DA-FRAC and LAC. \rac{} consists
of three key components: (1) a dedicated GNN-based architecture to
predict generator commitments and active transmission lines; (2) the
use of the GNN-based architecture and MC-dropout to associate a
confidence value to each commitment prediction; and (3) a
feasibility-restoration procedure that transforms the prediction into
a high-quality feasible solution in polynomial time. Experimental
results on the French transmission system and the DA-FRAC and LAC
formulations used by MISO indicate that \rac{} can provide significant
speedups in solving for negligible loss in solution quality. The
\rac{} framework sheds light on the use of confidence measurement and
uncertainty quantification in ML approaches to power systems. Future
research will explore how to further improve the confidence
calibration and the architectural design of the GNNs.  
The \rac{}
framework can be generalized to the stochastic optimization versions
of the FRAC and the LAC.
Additionally, this framework can be further enhanced to address network topology changes, thereby extending its applicability to accelerate optimal transmission planning or generalize transmission contingencies.

\section*{Acknowledgments}
This research is partly supported by NSF under Award Number 2007095
and 2112533, and ARPA-E, U.S. Department of Energy under Award Number
DE-AR0001280.

\bibliographystyle{IEEEtran} 
\bibliography{refs}

\newpage
\appendices

\section{Details on the Learning Architectures}
\label{appdx:arch_detail}
    
\begin{table}[!t]
        \centering
        \resizebox{0.95\columnwidth}{!}{
        \begin{tabular}{@{}ccccc@{}}
        \toprule
        Problem                  & Component                  & Encoder Name   & Input Dim. & Output Dim.\\
        \midrule
        \multirow{4}{*}{DA-FRAC} & \multirow{3}{*}{Generator} & Continuous generator encoder & 249  & 176 \\
                                 &                            & Binary generator encoder     & 55   & 32 \\
                                 &                            & Energy generator encoder     & 480  & 16 \\\cmidrule{2-5}
                                 & Load                       & Load encoder                 & 24   & 32 \\
        \midrule
        \multirow{4}{*}{LAC}     & \multirow{3}{*}{Generator} & Continuous generator encoder     & 141 & 176 \\
                                 &                            & Binary generator encoder         & 43  & 32 \\
                                 &                            & Energy generator encoder         & 240 & 16 \\\cmidrule{2-5}
                                 & Load                       & Load encoder                     & 12  & 32 \\
        \bottomrule
        \end{tabular}
        }
\caption{Input/output dimensions of the generator and load encoders for DA-FRAC and LAC} 
\label{tab:input_dim}
\end{table}
        
\paragraph{Encoders}

The architecture proposed in Section~\ref{sec:architecture} is
described here in detail.  The generator encoder $\text{Enc}_g$ can be
separated into three components; binary encoder, continuous encoder,
and energy encoder.  The binary encoder only encodes the binary type
input configurations to the latent representations. Similarly, the
continuous encoder encodes the continuous input configuration
parameters. The energy encoder encodes the energy step width and price
to the latent representations. The reasons for having three separable
encoders for the generator are as follows. First, it allows for a
reduction in the number of trainable parameters associated with the
MLP structures. Indeed, given $n$ dimensional input features, the
number of trainable parameters between two consecutive fully-connected
layers is $n^2$, which can be reduced to $3\left(\frac{n}{3}\right)^2$
when the input features are equally separated to three
encoders. Second, the input configurations for the energy step width
and price are quite sparse; most of the values are zero, meaning those
have less information. Thus, the energy encoder is designed to output
a smaller dimensional latent feature than other encoders. Third,
trainable parameters for the binary features and continuous features
can be affected by each other, thus for better normalization of the
weight parameters associated with fully-connected layers, the encoders
for binary and continuous features should be separated.
Table~\ref{tab:input_dim} illustrates the dimensions for each
binary/continuous/energy input configuration parameters per generator
as well as for a load unit. Note that all the trainable parameters of
the encoders are shared for all generators and load units.  All the
encoders are comprised of MLP structures. A continuous generator
encoder has $3$ fully-connected layers, whereas others have $2$
fully-connected layers. All the fully-connected layers are followed by
one-dimensional batch normalization \cite{ioffe2015batch} and a ReLU activation.
The number of hidden nodes for all the encoders are the average of the
input and output dimensions.  Specifically, the nodes for the binary
generator encoder are $55-43-32$ and $43-37-32$ for the DA-FRAC and LAC, respectively. 
Here, the first and last numbers represent, respectively, input and output
feature dimensions.  The number of nodes for the energy generator
encoder are $480-248-16$, and $240-128-16$ for the DA-FRAC and LAC, respectively.  
The number of nodes for the continuous generator encoder are
$249-212-212-176$ and $141-158-158-176$ for the DA-FRAC and LAC, respectively.
The output dimension of the encoders is $176+32+32+16=256$ for both
cases.
        
\paragraph{Graph Neural Networks}

The \texttt{GCN-1} architecture uses one GCN layer where the weight
has $256\times256$ trainable parameters.  \texttt{GCN-5} contains five
GCN layers that also output a 256-dimensional vector.  \texttt{SIGN}
has $5$ GCN layers with the normalized adjacency matrix and $3$ GCN
layers with the diffusion operator. Each GCN layer outputs a
64-dimensional vector, thus the associated trainable parameters form a
$256\times64$ dimensional matrix. The eight output vectors of
individual GCN layers are concatenated and processed by two-layered
MLP.  The number of nodes for this MLP is $8\times64-384-256$, and the
MLP produces a 256-dimensional vector for each bus of the network
structure.
        
\paragraph{Decoders}

The commitment decoder is also an MLP structure with bus
configurations of the form $256-210-210-24$, and $256-201-201-12$ for 
the DA-FRAC and LAC, respectively. Each layer is followed by a ReLU
activation. A dropout layer with a dropout probability of $0.5$ is
attached at the penultimate layer and is used to measure the
confidence of the predictions. Note that the output dimension
corresponds to the number of time steps. The sigmoid is attached to
the output of the commitment decoder to furnish the probability of
committing each generator. The number of nodes for the transmission line
decoder are the same as those for the commitment decoder. Each layer
is also followed by a ReLU activation.
        
\paragraph{Total Number of Trainable Parameters}
        
\begin{table}[t!]
\centering
\resizebox{0.95\columnwidth}{!}{
\begin{tabular}{@{}c|cc@{}}
\toprule
Problem                  & Model        & \# Trainable Parameters \\
\midrule
\multirow{5}{*}{DA-FRAC} & \texttt{MLP}          &  859,143,472 \\
                         & \texttt{Encoders+MLP} &  133,980,184 \\
                         & \texttt{GCN-1}        &  1,159,896 \\
                         & \texttt{GCN-5}        &  1,423,064 \\
                         & \texttt{SIGN}         &  2,177,369 \\
\midrule
\multirow{5}{*}{LAC}     & \texttt{MLP}          &  460,670,104 \\
                         & \texttt{Encoders+MLP} &  122,039,458 \\
                         & \texttt{GCN-1}        &  818,733 \\
                         & \texttt{GCN-5}        &  1,081,901 \\
                         & \texttt{SIGN}         &  1,871,078 \\
\bottomrule
\end{tabular}
}
\caption{The Number of Trainable Parameters of the Models for DA-FRAC and LAC.} 
\label{tab:trainparams}
\end{table}
        
The number of total trainable parameters of the proposed GNN-based
architectures and the other two deep baselines are summarized in
Table~\ref{tab:trainparams}.  As shown in the table, the number of
trainable parameters of the GNN-based architectures is much smaller
than that of the dense baselines. This is critical to scale to realistic
networks. 
        
\section{Details on Training}
\label{appdx:training}

Denote by $\mathbf{p}$ the input configuration vector that contains
the generator input features and load input features for all
generators and loads. Denote by $\mathbf{x}^*\in\{0,1\}$ the
ground-truth optimal commitment decisions that are obtained by solving
the optimization problems.  The optimal commitment decisions
$\mathbf{x}^*$ is a $|\mathcal{G}|\!\times\!|\mathcal{T}|$ dimensional
matrix in which an element $x_{gt}^*$ corresponds to the optimal
commitment decision for the generator $g$ at time step $t$.  For each
generator, recall that there are some must-run commitments, i.e.,
$x_{gt}^{\text{prev}}=1$ that do not need to be predicted as they are
determined by previous commitment decisions.

For instance $i$ and the commitment estimation $\tilde{\mathbf{x}}$
produced by the ML model, the commitment loss function $L_{c}$
minimized during training is the binary cross entropy between
$\tilde{\mathbf{x}}$ and $\mathbf{x}^*$ defined as
\begin{equation*}
    L_c^{(i)} = -\sum_{\substack{g\in\mathcal{G},t\in\mathcal{T}\\ | x_{gt}^{\text{prev}}=0}} x_{gt}^{*(i)}\log \tilde{x}_{gt}^{(i)} +(1-x_{gt}^{*(i)}) \log (1-\tilde{x}_{gt}^{(i)}).
\end{equation*}
Similarly, if the ground-truth transmission constraint is
given by $\mathbf{a}^*$ where 1 represents being active, 
the transmission loss function is defined as
\begin{equation*}
L_t^{(i)} = -\sum_{t\in\mathcal{L}'} a_{t}^{*(i)}\log \tilde{a}_{t}^{(i)} +(1-a_{t}^{*(i)}) \log (1-\tilde{a}_{t}^{(i)}),
\end{equation*}
where $\mathcal{L}'$ comprises of active transmission constraints and 
some samples of non active transmission constraints, called negative sampling
\cite{mikolov2013distributed}, which is widely used in natural language processing.
The use of $\mathcal{L}'$ instead of whole transmission set $\mathcal{L}$
is due to the sparsity of active transmissions and maintaining the balance between two classes.

Given $N$ instances in the training dataset, the loss function can be defined as
\begin{equation}
\label{eq:loss}
L = \frac{1}{N}\sum_{i=1}^N\left( L_c^{(i)} +\lambda L_t^{(i)} \right),
\end{equation}
where $\lambda$ is a hyper-parameter to balance between the two loss
terms. By minimizing the loss function~\eqref{eq:loss} with respect to
the trainable parameters, the estimations for commitment decisions and
active transmission constraints are obtained simultaneously in an
end-to-end manner. In the experiments, $\lambda$ is set to $1$ for
training.
    
\section{Details on the Diffusion Operator}
\label{appdx:diffusion_operator}

The generalized graph diffusion matrix used in this study is defined
as follows \cite{klicpera2019diffusion}:
\begin{equation}
\label{eq:diffusion_mat}
\mathbf{S}=\sum_{i=0}^\infty \theta_k\tilde{\mathbf{A}}^k,
\end{equation}
with the weighting coefficients $\theta_k$ and the normalized
adjacency matrix. The weighting coefficients should ensure that
$\sum_{k=0}^\infty \theta_k =1$ and $\theta_k \in [0,1]$. The
personalized PageRank (PPR) based diffusion operator exploits
$\theta_k=\alpha(1-\alpha)^k$ with teleport probability
$\alpha\in(0,1)$ to satisfy the aforementioned requirements.  Because
of the boundedness of the eigenvalues of $\tilde{\mathbf{A}}$ and the
property of the geometric series, the closed form of the PPR-based
diffusion operator can be obtained as
\begin{equation}
\label{eq:diffusion_mat2}
\mathbf{S} = \sum_{i=0}^\infty \alpha (1-\alpha)^k\tilde{\mathbf{A}}^k = \alpha \left( \mathbf{I}_N + (\alpha-1)\tilde{\mathbf{A}} \right) ^{-1},
\end{equation}
where $\mathbf{I}_N$ is the $N\times N$ identity matrix. Note that
$\mathbf{S}$ is now a dense matrix, and should be sparsified to
represent the spacial locality using the top-$k$ parameter. The sparse
diffusion operator with the top-$k$ parameter distills the $k$ highest
entries per column and other entries are set to zero.
In the experiment, top-$k$ parameter is set to $256$.
        
\section{Performance Results on Active Transmission Line Constraints}
\label{appdx:active_transmission}
    
\begin{table*}[!th]
\centering
\addtolength\tabcolsep{-2pt}
\begin{tabular}{l|ccccc}
\toprule
Month & Acc. & AUPR & AUROC & Pos. Ratio & \#Pos.\\ \midrule
\multicolumn{6}{c}{DA-FRAC}\\ \midrule
Feb. & 0.9982(0.0012) & 0.3462(0.1619) & 0.9548(0.0586) & 0.0008(0.0003) & 350.4200(142.2894) \\
Mar. & 0.9994(0.0004) & 0.2494(0.1485) & 0.9515(0.0721) & 0.0005(0.0004) & 232.3700(173.6947) \\
Apr. & 0.9999(0.0001) & 0.5571(0.1935) & 0.9901(0.0111) & 0.0001(0.0000) & 45.2240(16.8886) \\
May & 0.9999(0.0000) & 0.5571(0.1935) & 0.9901(0.0111) & 0.0001(0.0000) & 45.2240(16.8886) \\
June & 1.0000(0.0000) & 0.9223(0.0652) & 0.9985(0.0047) & 0.0001(0.0000) & 42.4920(12.1598) \\
July & 1.0000(0.0000) & 0.7566(0.1136) & 0.9875(0.0132) & 0.0001(0.0000) & 39.1666(10.8016) \\
Aug. & 0.9999(0.0000) & 0.8103(0.1137) & 0.9887(0.0196) & 0.0001(0.0000) & 47.5480(15.3757) \\
Sep. & 1.0000(0.0000) & 0.8048(0.1365) & 0.9867(0.0300) & 0.0001(0.0000) & 24.3260(9.7112) \\
Oct. & 0.9999(0.0002) & 0.4794(0.3129) & 0.8826(0.1153) & 0.0001(0.0002) & 47.2500(69.4714) \\
Nov. & 0.9994(0.0006) & 0.2611(0.1493) & 0.9317(0.0496) & 0.0005(0.0003) & 211.7900(140.1693) \\
Dec. & 0.9991(0.0006) & 0.4230(0.1056) & 0.9820(0.0115) & 0.0007(0.0004) & 304.0460(160.2599) \\ \midrule
Total & 0.9996(0.0007) & 0.5400(0.2857) & 0.9672(0.0592) & 0.0003(0.0004) & 125.2884(152.1879) \\ \midrule\midrule
\multicolumn{6}{c}{LAC}\\ \midrule
Feb. & 0.9989(0.0006) & 0.5621(0.1656) & 0.9703(0.0583) & 0.0008(0.0003) & 164.6520(71.6983) \\
Mar. & 0.9994(0.0006) & 0.5436(0.1899) & 0.9948(0.0336) & 0.0005(0.0007) & 107.5860(143.1241) \\
Apr. & 0.9999(0.0001) & 0.5026(0.3421) & 0.9985(0.0063) & 0.0001(0.0001) & 19.6750(15.7187) \\
May  & 0.9999(0.0000) & 0.5883(0.3719) & 0.9971(0.0129) & 0.0001(0.0001) & 26.6875(20.1051) \\
June & 1.0000(0.0000) & 0.8631(0.2637) & 0.9476(0.1336) & 0.0001(0.0001) & 27.2336(21.2843) \\
July & 0.9999(0.0000) & 0.6861(0.3249) & 0.9985(0.0093) & 0.0001(0.0001) & 28.1809(24.4316) \\
Aug. & 0.9999(0.0000) & 0.7415(0.3088) & 0.9946(0.0319) & 0.0002(0.0001) & 33.9729(27.2578) \\
Sep. & 1.0000(0.0000) & 0.7104(0.3599) & 0.9881(0.0637) & 0.0001(0.0001) & 21.4605(20.8516) \\
Oct. & 0.9999(0.0003) & 0.4614(0.3618) & 0.9598(0.0651) & 0.0002(0.0003) & 36.5758(55.1402) \\
Nov. & 0.9994(0.0007) & 0.5904(0.2037) & 0.9611(0.0611) & 0.0005(0.0005) & 105.9457(110.6117) \\
Dec. & 0.9994(0.0004) & 0.6168(0.1343) & 0.9936(0.0083) & 0.0007(0.0003) & 143.2720(69.7454) \\ \midrule
Total & 0.9997(0.0005) & 0.6178(0.2993) & 0.9818(0.0594) & 0.0003(0.0004) & 70.9388(88.5095) \\
\bottomrule
\end{tabular}
\caption{Performance Results on Active Transmission Constraints in
  DA-FRAC and LAC. Accuracy (Acc.), area under precision-recall curve (AUPR), and
  area under receiver operating characteristic curve (AUROC) are
  presented as performance measures. The standard deviations are shown
  in parenthesis. Pos. Ratio represents the ratio of the positive
  labels (active transmissions) and \#Pos is the averaged number of
  active transmissions in each test instance. The results are obtained
  from 500 instances per month.}
\label{table:active_transmission}
\end{table*}

The classification performance for active transmission constraints is
reported in Table~\ref{table:active_transmission} for RAC
optimizations. The results are based on \texttt{SIGN} that is also used 
for reporting the commitment decision
performance and speedup results.  The results are calculated using 500
test instances per each test month from February to December in 2018.
More transmission lines are congested in the winter, leading to a
higher number of active transmission lines compared to the summer
season. This also makes it harder to predict the active transmission
constraints in the winter, mirroring the results for commitment
decisions in Fig~\ref{img:month_performance}.
    
The proportion of the active transmission constraints is quite
unbalanced, which is around $0.03\%$ for both RACs.  Accuracy (Acc.)
and AUROC values are relatively high, but those are not adequate
measure to denote the performance for imbalanced classification
problems. Instead, the area under precision recall curve (AUPR) is the
best choice as a measure of classification performance. The baseline
AUPR value in this case with a randomized classifier is $0.0003$,
which is the same as the ratio of the positive labels.  Recall that
the prediction of active transmission is less sensitive than
commitment decisions. Note that \rac{} could adds false negative predictions in
the iterative optimization procedure. These false negative lines do
not affect the feasibility of the optimization model, but slow the
solver down slightly.

\section{Detailed Results on the Optimization Speedups}
\label{appdx:optimization_speedup}

\begin{table*}[!t]
\caption{Speedup Ablation Study for DA-FRAC. The speedup performances of the proposed approach are denoted in bold.}
\centering
\addtolength\tabcolsep{-2pt}
\begin{tabular}{cc|cc|c V{2} cccc}
\toprule
ADD$^{1}$ & \#Added & FIX$^{2}$ & Ratio(\%) & FR$^{3}$ & Feasibility(\%) & Opt.Gap(\%) & Times(s) & Speedup \\ \midrule
\xmark                  & -                       & \xmark                  & -   & \xmark                  & 100.00 & 0.00 &126.06 & 1.0000$\times$        \\\midrule
\cmark                  & 147.12                  & \xmark                  & -   & \xmark                  & 100.00 & 0.00 & 61.15 & 2.0581$\times$        \\\midrule
\multirow{5}{*}{\xmark} & \multirow{5}{*}{-}      & \multirow{5}{*}{\cmark} & 30  & \multirow{5}{*}{\xmark} & 100.00 & 0.01 & 94.56 & 1.3595$\times$        \\
                        &                         &                         & 50  &                         & 100.00 & 0.02 & 91.19 & 1.4074$\times$        \\
                        &                         &                         & 70  &                         & 100.00 & 0.06 & 83.99 & 1.5261$\times$        \\
                        &                         &                         & 90  &                         &  89.61 & 0.07 & 63.75 & 1.8471$\times$        \\
                        &                         &                         & 100 &                         &  31.17 & 0.12 & 32.19 & 2.4268$\times$        \\\midrule
\multirow{5}{*}{\cmark} & \multirow{5}{*}{147.12} & \multirow{5}{*}{\cmark} & 30  & \multirow{5}{*}{\xmark} & 100.00 & 0.01 & 48.93 & 2.6267$\times$        \\
                        &                         &                         & 50  &                         & 100.00 & 0.02 & 47.56 & 2.6938$\times$        \\
                        &                         &                         & 70  &                         & 100.00 & 0.06 & 45.07 & 2.8405$\times$        \\
                        &                         &                         & 90  &                         &  89.61 & 0.07 & 35.82 & 3.2800$\times$        \\
                        &                         &                         & 100 &                         &  31.17 & 0.12 & 19.19 & 4.0674$\times$        \\\midrule
\multirow{5}{*}{\cmark} & \multirow{5}{*}{147.12} & \multirow{5}{*}{\cmark} & 30  & \multirow{5}{*}{\cmark} & 100.00 & 0.01 & 48.40 & \textbf{2.6571}$\times$        \\
                        &                         &                         & 50  &                         & 100.00 & 0.02 & 47.36 & \textbf{2.7078}$\times$        \\
                        &                         &                         & 70  &                         & 100.00 & 0.06 & 45.03 & \textbf{2.8436}$\times$        \\
                        &                         &                         & 90  &                         & 100.00 & 0.16 & 38.81 & \textbf{3.3135}$\times$        \\
                        &                         &                         & 100 &                         & 100.00 & 0.77 & 31.56 & \textbf{4.1160}$\times$        \\\bottomrule
\end{tabular}
\\
1) ADD: whether to add active transmission constraints prediction at the beginning of the iterative optimization. \\
2) FIX: whether to fix highly confident commitment decisions to their predictions. \\
3) FR: whether to use feasibility restoration.
\label{table:speedup_frac}
\end{table*}
    
\begin{table*}[!t]
\caption{Speedup Ablation Study for LAC. The speedup performances of the proposed approach are denoted in bold.}
\centering
\addtolength\tabcolsep{-2pt}
\begin{tabular}{cc|cc|c V{2} cccc}
\toprule
ADD & \#Added & FIX & Ratio(\%) & FR & Feasibility(\%) & Opt.Gap(\%) & Times(s) & Speedup \\ \midrule
\xmark                  & -                       & \xmark                  & -   & \xmark                  & 100.00 & 0.00 & 42.71 & 1.0000$\times$        \\\midrule
\cmark                  & 119.94                  & \xmark                  & -   & \xmark                  & 100.00 & 0.00 & 29.11 & 1.4472$\times$        \\\midrule
\multirow{5}{*}{\xmark} & \multirow{5}{*}{-}      & \multirow{5}{*}{\cmark} & 30  & \multirow{5}{*}{\xmark} & 100.00 & 0.00 & 35.97 & 1.2143$\times$        \\
                        &                         &                         & 50  &                         & 100.00 & 0.01 & 34.79 & 1.2577$\times$        \\
                        &                         &                         & 70  &                         &  99.53 & 0.01 & 32.93 & 1.3293$\times$        \\
                        &                         &                         & 90  &                         &  98.58 & 0.01 & 30.97 & 1.4030$\times$        \\
                        &                         &                         & 100 &                         &  84.91 & 0.09 & 25.96 & 1.6445$\times$        \\\midrule
\multirow{5}{*}{\cmark} & \multirow{5}{*}{119.94} & \multirow{5}{*}{\cmark} & 30  & \multirow{5}{*}{\xmark} & 100.00 & 0.00 & 26.44 & 1.6012$\times$        \\
                        &                         &                         & 50  &                         & 100.00 & 0.00 & 25.08 & 1.6987$\times$        \\
                        &                         &                         & 70  &                         &  99.53 & 0.01 & 25.04 & 1.6844$\times$        \\
                        &                         &                         & 90  &                         &  98.59 & 0.01 & 21.87 & 1.9460$\times$        \\
                        &                         &                         & 100 &                         &  84.51 & 0.09 & 21.59 & 1.8772$\times$        \\\midrule
\multirow{5}{*}{\cmark} & \multirow{5}{*}{119.94} & \multirow{5}{*}{\cmark} & 30  & \multirow{5}{*}{\cmark} & 100.00 & 0.00 & 25.45 & \textbf{1.6719}$\times$        \\
                        &                         &                         & 50  &                         & 100.00 & 0.00 & 24.82 & \textbf{1.7086}$\times$        \\
                        &                         &                         & 70  &                         & 100.00 & 0.01 & 23.67 & \textbf{1.8018}$\times$        \\
                        &                         &                         & 90  &                         & 100.00 & 0.01 & 22.73 & \textbf{1.8781}$\times$        \\
                        &                         &                         & 100 &                         & 100.00 & 0.09 & 20.94 & \textbf{2.0585}$\times$        \\\bottomrule
\end{tabular}
\\
\label{table:speedup_lac}
\end{table*}

This Appendix analyzes the contributions of each component of \rac{}
to the speedups reported in Section \ref{ssec:speedup}. It shows
that the learning benefits for commitments and active constraints
are cumulative. It also shows that feasibility restoration is critical
in practice.

The experiments consider four settings for the confident subset: 30\%,
50\%, 70\%, or 90\% of the highly confident commitment predictions.
For example, in the 30\% confident subset, 30\% of highly confident
commitments will be fixed to their predictions. For the active
transmission constraints, \rac{} adds to the optimization algorithm
all constraints whose probability of being active\footnote{Active
  constraints are those which are active or violated.} is greater than
$0.1$ are added to the beginning of the iterative optimization
procedure. This is a conservative threshold but it is counteracted by
the fact that \rac{} only add at most $400$ constraints. 
        
The results of speedup with some components ablated are depicted in
Table~\ref{table:speedup_frac} and Table~\ref{table:speedup_lac} for
the DA-FRAC and the LAC. In the experiments, 20 instances per month
are drawn at random from the monthly test dataset from February to
December.  The optimality gap (represented as Opt. Gap in tables) is
calculated as the geometric mean of relative differences in objective
values between the baseline and reduced problem divided by the
baseline objective value on the feasible instances. To treat some
negative values when calculating the geometric mean, 0.1\% is used as
a shift value.  Also, the time and speedup columns in the tables
represent the geometric means of elapsed times for solving the
optimization instances and their fractions based on the baseline case.
        
The baseline (denoted as `ADD', `FIX', and `FR' with \xmark{}) is the
iterative method \cite{xavier2019transmission} with $k=15$ as
suggested in \cite{xavier2021learning}. According to
\cite{barry2022risk}, at MISO, the DA-FRAC and LAC must be solved
within 40, 10 minutes.  The elapsed times of the baseline iterative
method for DA-FRAC and LAC is $126.06$ and $42.71$ seconds,
respectively, which are around 14-20 times faster than the time
limits, meaning that the baseline is quite competitive.
        
The second row of the result of the tables (denoted as `ADD':\cmark{},
`FIX':\xmark{}, `FR':\xmark{}) shows the computational results when
only the active constraint predictions are considered. It is important
to note that these predictions do not affect the feasibility of the
optimization, since the algorithm will add active constraints as
needed. The predictions only affect the running times by reducing the
number of iterations. The addition of transmission constraints results
in speedups of $2.0581\!\times$ and $1.4472\!\times$ for the DA-FRAC
or the LAC, respectively.  The total number of the transmission
constraints is $2\!\times\!8965\!\times\!24=430,320$ and
$2\!\times\!8965\!\times\!12=215,160$ for the DA-FRAC and the LAC,
(the $2$ comes from the min/max bounds).  The number of added transmission
constraints (denoted as `\#Added' in the tables) at the beginning of
the iterative method is relatively small; $147.12$ or $119.94$ for the
DA-FRAC and the LAC in average.  By adding around $0.02$-$0.06\%$ of
the transmission constraints at the beginning of the iterative
optimization procedure, speedups of around $40$-$100\%$ can be
obtained over the baseline.

The next lines in the table reports result when only generator
commitment predictions are used (the FIX column), which is denoted as
`ADD':\xmark{}, `FIX':\cmark{}, `FR':\xmark{} in the tables. 
Fixing $90\%$ of highly confident commitment predictions leads
to speedups of $1.8471\times$, $1.4030\times$ for
the DA-FRAC and the LAC. However, about 10\% of the instances
become infeasible in the DA-FRAC when doing so. The infeasibility issue
is especially severe when fixing all the predictions. 

The next results consider the case where both the commitment
predictions and the active constraint predictions are used together.
This is represented by (`ADD':\cmark{}, `FIX':\cmark{},
`FR':\xmark{}). The main result here is that the speedups are cumulative,
especially for the DA-FRAC. The infeasibilies are still present obviously.

The last set of lines adds the feasibility restoration to obtain the
\rac{} framework (denoted by (`ADD':\cmark{}, `FIX':\cmark{} and
`FR':\cmark{}) in the table). Feasibility restoration marginally
increases the optimality gaps in general and the \rac{} framework
speeds up the optimization by factor of about 4 and 2 for the DA-FRAC
and the LAC.

\end{document}